\documentclass[10pt,twocolumn,letterpaper]{article}

\usepackage{cvpr}              % To produce the CAMERA-READY version
\usepackage[accsupp]{axessibility} 
\definecolor{cvprblue}{rgb}{0.21,0.49,0.74}
\usepackage[pagebackref,breaklinks,colorlinks,allcolors=cvprblue]{hyperref}

\usepackage{multirow}
\usepackage{tabularx}
\usepackage{xcolor}         % colors
\usepackage{subcaption}
\usepackage{multirow}
\usepackage{pifont}
\usepackage[table]{xcolor}

\def\paperID{} % *** Enter the Paper ID here
\def\confName{CVPR}
\def\confYear{2026}

\title{Mitigating the ID–OOD Tradeoff in Open-Set Test-Time Adaptation}

\author{Wenjie Zhao$^{1}$ \hspace{0.5cm} Jia Li$^{1}$ \hspace{0.5cm} Xin Dong$^{2}$ \hspace{0.5cm} Yapeng Tian$^{1}$ \hspace{0.5cm} Yu Xiang$^{1}$ \hspace{0.5cm} Yunhui Guo$^{1}$\\
$^{1}$University of Texas at Dallas \hspace{0.5cm} $^{2}$ NVIDIA\\
$^{1}${\tt\small \{wenjie.zhao, yunhui.guo\}@utdallas.edu}
}

\begin{document}
\maketitle
\begin{abstract}
Open-set test-time adaptation (OSTTA) addresses the challenge of adapting models to new environments where out-of-distribution (OOD) samples coexist with in-distribution (ID) samples affected by distribution shifts. In such settings, covariate shift—for example, changes in weather conditions such as snow—can alter ID samples, reducing model reliability. Consequently, models must not only correctly classify covariate-shifted ID (csID) samples but also effectively reject covariate-shifted OOD (csOOD) samples. Entropy minimization is a common strategy in test-time adaptation to maintain ID performance under distribution shifts, while entropy maximization is widely applied to enhance OOD detection. Several studies have sought to combine these objectives to tackle the challenges of OSTTA. However, the intrinsic conflict between entropy minimization and maximization inevitably leads to a trade-off between csID classification and csOOD detection. In this paper, we first analyze the limitations of entropy maximization in OSTTA and then introduce an angular loss to regulate feature norm magnitudes, along with a feature-norm loss to suppress csOOD logits, thereby improving OOD detection. These objectives form ROSETTA, a \underline{r}obust \underline{o}pen-\underline{se}t \underline{t}est-\underline{t}ime \underline{a}daptation method. Our method achieves strong OOD detection while maintaining high ID classification performance on CIFAR-10-C, CIFAR-100-C, Tiny-ImageNet-C and ImageNet-C. Furthermore, experiments on the Cityscapes validate the method’s effectiveness in real-world semantic segmentation, and results on the HAC dataset demonstrate its applicability across different open-set TTA setups.
\end{abstract}

\section{Introduction}
\label{sec:intro}

\begin{figure}[t]
    \centering
    \scriptsize
    \begin{subfigure}{0.38\columnwidth}
        \centering
        \includegraphics[width=\linewidth]{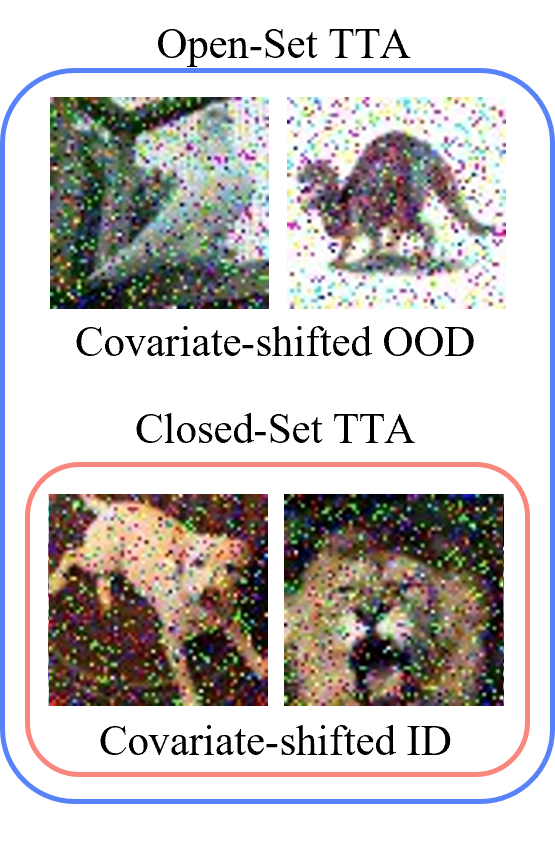}
        \caption{}
        \label{fig:csidood}
    \end{subfigure}
    \hfill
    \begin{subfigure}{0.58\columnwidth}
        \centering
        \includegraphics[width=\linewidth]{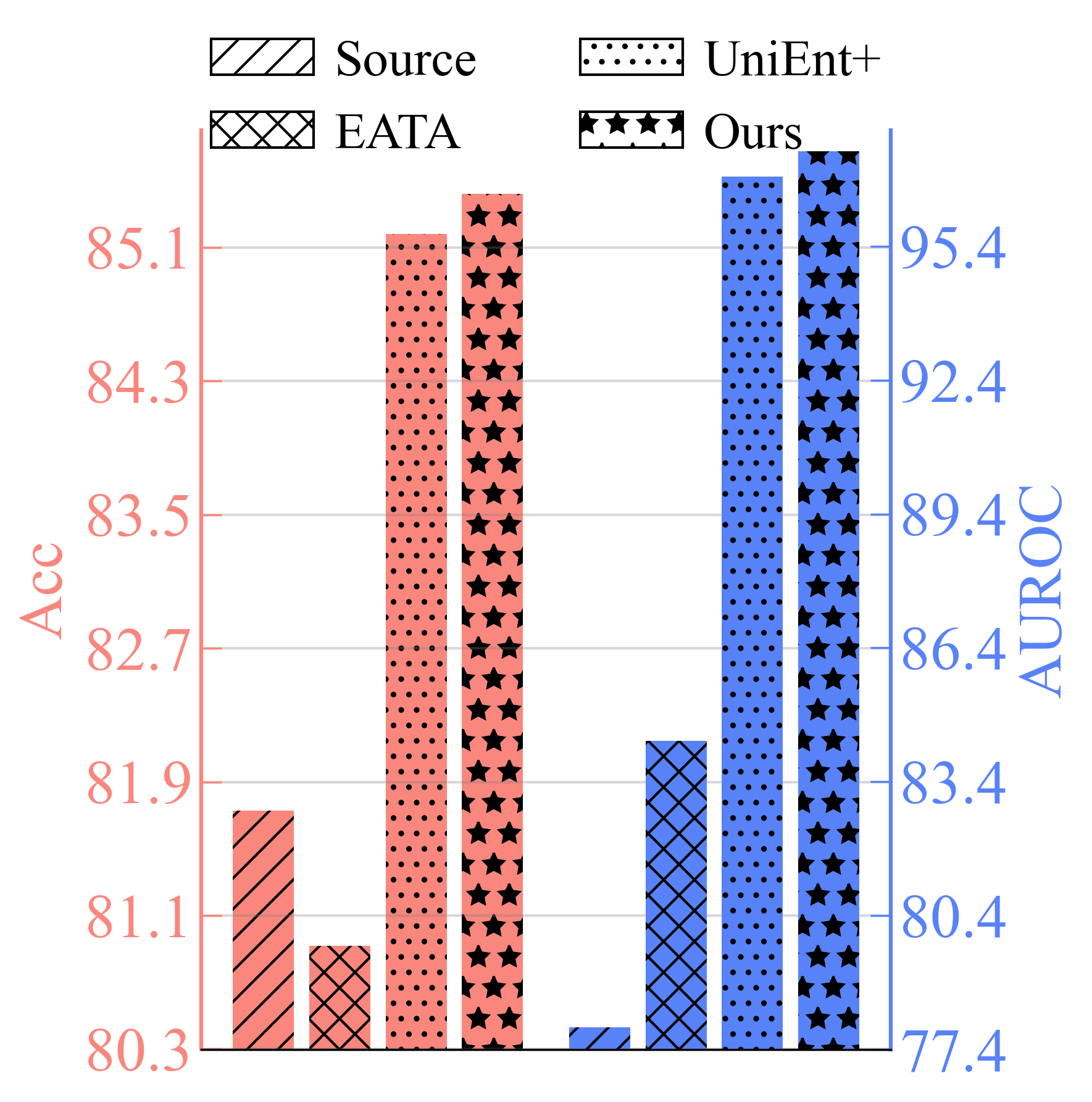}
        \caption{}
        \label{fig:compare}
    \end{subfigure}

    \caption{(a) Examples of csID and csOOD samples. Open-set TTA adapts the model to handle both simultaneously. (b) 
    % Recent state-of-the-art methods \cite{gao2024unient} usually excel at either ID classification (Acc) or OOD detection (AUROC), but not both. 
    Comparison with recent state-of-the-art methods.
    Our method achieves strong performance on both tasks.}

    \label{fig:two_images}
\end{figure}

Covariate shift, which occurs when environmental or stylistic variations change the input distribution without altering the task itself, is frequently encountered in real-world applications. Models trained on clean data often exhibit performance degradation when confronted 
with covariate-shifted inputs \cite{hendrycks2019benchmarking}.
%recht2019imagenet, geirhos2018generalisation, wang2020tent
To mitigate this issue, test-time adaptation (TTA) methods have been developed for models to adapt to covariate-shifted data during inference \cite{zhou2022domain, yang2023fsood}. These methods assume a closed-set setting where only covariate-shifted in-distribution (csID) samples are present. In practice, models are frequently exposed to out-of-distribution (OOD) inputs from novel and unknown classes, which should be identified as OOD \cite{liang2017enhancing,Liu_2025_CVPR}.
% For reliable deployment, such samples should be identified as OOD. 
% When csOOD samples appear alongside csID samples due to environmental noise, this scenario is referred to as open-set TTA (OSTTA). 
When covariate shifts introduce csOOD samples alongside csID samples, this scenario is referred to as open-set TTA (OSTTA).
The goal of OSTTA is to classify csID samples affected by covariate shift, while effectively rejecting csOOD samples (Fig.~\ref{fig:csidood}). 

 % gao2024unient, dong2025towards
Several studies have recognized the challenges of OSTTA and have attempted to extend TTA to open-set settings \cite{li2023robustness, lee2023towards, yuan2023robust, yu2024stamp}. A key challenge in OSTTA is distinguishing OOD samples from ID samples. Entropy maximization has been widely adopted in the OOD detection literature to encourage models to assign low confidence to OOD samples, thereby enhancing their ability to discriminate between OOD and ID samples \cite{dong2025towards, gao2024unient, chan2021entropy, de2025deep, macedo2021enhanced}. Given its success and seamless integration, entropy maximization has become a key component in state-of-the-art open-set TTA methods such as UniEnt \cite{gao2024unient} and AEO \cite{dong2025towards}.
% \begin{wrapfigure}{r}{0.5\linewidth}  % r 表示右侧, 宽度为整页的一半
%     \centering
%     \scriptsize
%     % \vspace{-10pt} % 上方间距，可微调
%     \begin{minipage}{\linewidth}
%         \begin{subfigure}{0.38\linewidth}
%             \centering
%             \includegraphics[width=\linewidth]{image/csidood2.png}
%             \caption{}
%             \label{fig:csidood}
%         \end{subfigure}
%         \hfill
%         \begin{subfigure}{0.585\linewidth}
%             \centering
%             \includegraphics[width=\linewidth]{image/compare.png}
%             \caption{}
%             \label{fig:compare}
%         \end{subfigure}
%     \end{minipage}
%     \caption{a) Examples of covariate-shifted ID and OOD samples. Open-set TTA aims to adapt the model to handle both csID and csOOD samples simultaneously. b) Recent state-of-the-art methods \cite{gao2024unient} tend to excel at either ID classification (measured by Acc) or OOD detection (measured by AUROC), but not both. In contrast, our method achieves strong performance on both tasks.}
%     \vspace{-10pt} % 下方间距，可微调
% \end{wrapfigure}
In contrast, entropy minimization in TTA encourages the model to make confident predictions on corrupted inputs, helping to preserve classification accuracy. 
%In contrast, entropy maximization promotes uncertainty on OOD samples, which enhances the model’s ability to reject unknown inputs.
While both techniques are effective within their respective domains, we observe that combining them in OSTTA can create conflicting objectives, particularly because no OOD detector can reliably determine whether an input sample belongs to csID or csOOD. This conflict results in a trade-off between csID classification performance and csOOD detection. 
% As shown in Fig. \ref{fig:compare}, while UniEnt achieves stronger OOD detection, it suffers from reduced ID classification accuracy. In contrast, UniEnt+ yields better ID classification results but significantly weaker OOD detection performance.

In particular, minimizing entropy increases the feature norm across all inputs, weakening the separation between csID and csOOD samples. Furthermore, the inherent imperfections of the OOD detector, as analyzed in Sec.~\ref{sec3_3:inefficacy}, cause entropy maximization and minimization to conflict, ultimately harming overall performance. To improve csOOD detection while preserving csID classification, we propose Robust Open-Set Test-Time Adaptation (ROSETTA) based on our analysis, which decouples the optimization objectives for csOOD detection and csID classification. We introduce two complementary objectives: an angular loss that avoids feature norm inflation while enhancing csID alignment, and a feature norm loss that suppresses csOOD logits to better separate OOD samples from ID. ROSETTA formulates a unified optimization loss that explicitly disentangles csOOD detection and csID adaptation to achieve both objectives effectively. This design enables the model to achieve accurate csID classification and effective csOOD detection simultaneously, as illustrated in Fig.~\ref{fig:compare}.
%Our approach enhances model robustness, ensuring strong performance in the presence of label misassignments.

Our main contributions are summarized as follows:

\begin{itemize}
    % \item We investigate the limitations of entropy maximization in open-set TTA and identify how conflicting loss lead to interference between csID classification and csOOD detection.
    \item We analyze how the conflicting objectives of entropy minimization and entropy maximization interfere with each other in open-set TTA, ultimately degrading performance.
    
    \item We propose novel learning objectives that explicitly disentangle csID classification from csOOD detection. Our analysis shows that our design mitigates the conflict between the two tasks, leading to improved performance in both. 
    
    \item 
    We conduct comprehensive experiments on widely used robustness benchmarks, including CIFAR-10-C, CIFAR-100-C, Tiny-ImageNet-C, ImageNet-C, and Cityscapes-to-ACDC. Our method consistently outperforms the state-of-the-art open-set TTA method on both csID classification and csOOD detection. It further achieves strong results on semantic segmentation benchmarks, indicating its applicability beyond classification.
\end{itemize}

\vspace{-0.2cm}
\section{Related work}
\label{sec:related}
\vspace{-0.1cm}
\noindent\textbf{Test-time Adaptation} is increasingly studied as a solution to distribution shifts encountered during inference. TTA primarily focused on closed-set scenarios, where the source and target domains share identical class labels \citep{khurana2021sita, nado2020evaluating, niu2022efficient, wang2020tent}. For example, TENT \citep{wang2020tent} adapted models at test time by updating batch normalization statistics and optimizing entropy-based objectives. Subsequent methods such as CoTTA \citep{wang2022continual} and EATA \citep{niu2022efficient} extend the TTA framework by introducing mechanisms to mitigate catastrophic forgetting and reduce error accumulation. These methods incorporate teacher-student architectures, selective sampling, and anti-forgetting regularization to enhance model adaptability.

% \noindent\textbf{Full-spectrum OOD} (FS-OOD) detection focuses on handling both ID and OOD data in open-world scenarios \citep{lu2023likelihood, yang2023fsood}. This setting aims to distinguish between various types of OOD conditions, including covariate shift and semantic shift. Traditional OOD detection methods often struggle with these challenges, as they typically do not account for noisy or unpredictable conditions under practical conditions. 
% Recent open-set TTA methods tackle the FS-OOD detection problem by applying entropy maximization strategies \cite{gao2024unient, dong2025towards}. These approaches enhance model robustness under diverse and unpredictable input distributions in dynamic environments.

% \noindent\textbf{Full-spectrum OOD} (FS-OOD) detection extends the conventional OOD detection paradigm by considering both semantic shift and covariate shift \citep{yang2023fsood}. 
% Traditional OOD detection benchmarks typically assume that OOD samples differ semantically from the ID data, while csID samples are either ignored or tend to be misclassified as OOD due to appearance changes such as lighting or contrast, which can easily distort the detection scores. 
% This contradicts the goal of generalization in machine learning. The FS-OOD setting therefore provides a more realistic evaluation scenario where models are expected to distinguish between semantic-shifted OOD samples and csID samples.

\noindent\textbf{Full-spectrum OOD} (FS-OOD) detection extends the conventional OOD detection paradigm by considering both semantic shift and covariate shift \citep{yang2023fsood}. Most existing OOD detection methods focus primarily on semantic shift, assuming that OOD samples differ semantically from the ID data \cite{zhao2024segment,hendrycks2016baseline}. Under this setting, csID samples are often neglected or mistakenly treated as OOD because appearance changes such as lighting or contrast can distort detection scores. The FS-OOD setting therefore provides a more realistic evaluation scenario, where models are expected to distinguish semantic-shifted OOD samples from csID samples.

\noindent\textbf{Open-set Test-time Adaptation.}
Open-set TTA addresses scenarios where the incoming test data contain both csID and csOOD samples. Rather than focusing solely on detection, open-set TTA pursues a dual objective by distinguishing csOOD from csID while maintaining or improving task performance on csID under covariate shift. Recent approaches typically apply entropy maximization to strengthen OOD detection and entropy minimization to adapt on reliable csID samples \citep{gao2024unient, dong2025towards}. By integrating detection and adaptation within a unified framework, open-set TTA offers a more comprehensive view of model robustness. This setting highlights the practical need for models that can adapt to changing environments while remaining selective against unseen distributions.

% \noindent\textbf{OOD Detection with Feature Norm.} It has been widely observed that models trained on ID data tend to produce higher feature norms for ID samples compared to OOD samples \cite{dhamija2018reducing, yu2020out, chen2020norm, meng2021magface}. In biometrics, \cite{yu2020out} employed feature norm to reject unseen identities, and \cite{meng2021magface} further extended this idea to assess face image quality. \cite{chen2020norm, chen2021norm} also demonstrated that feature norm can help distinguish individuals from backgrounds, thereby improving person search. 
% % As research continues to explore the connection between feature norm and OOD detection,
% \cite{park2023understanding} offers a comprehensive theoretical analysis, showing that discriminative classifiers yield OOD features with smaller $l_1$-norms compared to those of ID samples.

\section{Background}

\subsection{Problem setup}

We denote the mini-batch of test samples arriving during adaptation at timestamp $t$ as $\mathcal{B}_t$ and the adapted model as $\mathcal{M}_\theta$. The mini-batch $\mathcal{B}_t$ contains csID samples from $\mathcal{D}_s = \{(\mathbf{x_i}, y_i)\}_{i=1}^{O}$, where \(y_i \in \mathcal{Y}_s = \{1, \ldots, K\}\) represents class labels from \(K\) classes,  and csOOD samples from $\mathcal{D}_t = \{(\mathbf{x_j}, y_j)\}_{j=1}^{P}$, where \(y_j \in \mathcal{Y}_t = \{1, \ldots, F\}\) represents class labels from \(F\) classes. The goal of open-set TTA is to correctly classify samples within $\mathcal{D}_s $ while rejecting those from $\mathcal{D}_t$.

\subsection{Notation}
Suppose the model $\mathcal{M}_\theta$ is a neural network consisting of \(l\) hidden layers. The feature vector of a given input $\mathbf{x}$ in the final hidden layer is computed as \(\mathbf{a}^{(l)}(\mathbf{x}) = \sigma \left( \mathbf{W}^{(l)^\mathrm{T}} \mathbf{a}^{(l-1)}(\mathbf{x}) \right)\), where \(\mathbf{W}^{(l)}\) is the weight matrix and \(\sigma\) is the activation function. For simplicity, we denote the feature extractor of the model  $\mathcal{M}_\theta$ as $ g_\theta$. A linear layer \(L\) with weight $\mathbf{W}^{(L)}$ is applied to produce the classification logits for each class, given by \(\psi(\mathbf{x}) = \mathbf{W}^{(L)^\mathrm{T}} \mathbf{a}^{(l)}(\mathbf{x})\). The model’s probability output is represented as $\mathcal{M}_\theta (x) = \textnormal{softmax}(\psi(\mathbf{x}))$.
 %put it later
% The \(L_1\)-norm of the feature vector \(\mathbf{a}^{(l)}\) in the \(l\)-th hidden layer is defined as \(\|\mathbf{a}^{(l)}\|_1 = \sum_{j=1}^{d_l} \left| a_j^{(l)} \right|\), where \(d_l\) is the dimensionality of the feature vector \(\mathbf{a}^{(l)}\) and \(a_j^{(l)}\) is the \(j^{\text{th}}\) element of \(\mathbf{a}^{(l)}\).
\subsection{Challenges in Distinguishing csID and csOOD for Open-Set TTA}
\label{sec3_3:inefficacy}
\begin{figure}[!t]
    \centering
    \includegraphics[width=\linewidth]{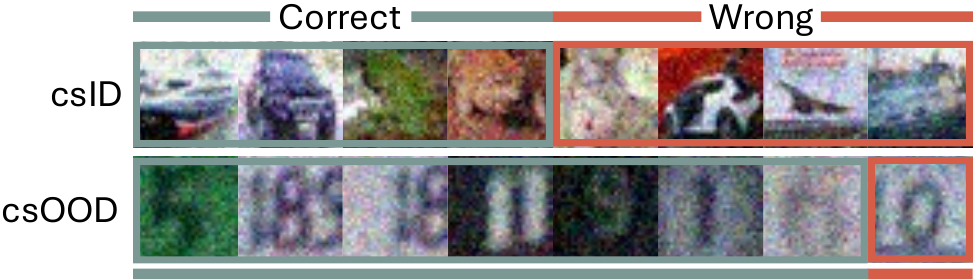}
    \caption{Examples of misclassification under the optimal entropy threshold. Green boxes represent correctly detected samples and red boxes represent misclassified ones, where csID images are often mistaken for csOOD.}
    \label{fig:detector}
\end{figure}

Recent open-set TTA methods generally follow a two-step process. They first apply a post-hoc OOD detection stage that separates each test batch into presumed ID and OOD samples based on confidence measures derived from the model’s outputs, such as energy or entropy.
For instance, UniEnt models the distribution of energy scores with a Gaussian mixture, whereas AEO applies a fixed entropy threshold. However, OOD detectors are inherently imperfect and inevitably misclassify a portion of samples. As shown in Fig.~\ref{fig:detector}, visually corrupted ID samples are frequently misclassified as csOOD. We evaluate several OOD detectors using WideResNet-40 across four datasets, where the accuracy reported in Table~\ref{tab:comparison_methods} denotes the proportion of samples correctly classified as csID or csOOD. As shown, even with the best possible threshold, a notable portion of samples remain misclassified. The best threshold is obtained by evaluating all possible threshold values to maximize accuracy, thus reflecting the upper limit of each score’s performance. Such misclassification leads to conflicting updates during adaptation, where entropy minimization is mistakenly applied to OOD samples and entropy maximization to ID ones. Such conflicts hurt both csID classification and csOOD detection.

This work aims to resolve the issue by decoupling these competing objectives, allowing both tasks to improve simultaneously.
% \begin{table}[htbp]
% \centering
% \scriptsize % 字体缩小一点（9pt）
% \renewcommand{\arraystretch}{0.8}
% \setlength{\tabcolsep}{5pt}
% \begin{tabular}{lcccc}
% \toprule
% Method & CIFAR-10-C & CIFAR-100-C & TinyImageNet-C & ImageNet-C \\
% \midrule
% GMM classifier        & 74.73 & 70.74 & 59.23 & 55.42 \\
% Entropy K-means       & 70.16 & 67.73 & 56.79 & 58.33 \\
% Energy K-means        & 73.92 & 65.86 & 56.78 & 55.00 \\
% Entropy (Best Thre.)  & 75.98 & 70.15 & 59.90 & 67.08 \\
% Energy (Best Thre.)   & 75.43 & 68.79 & 60.54 & 65.42\\
% \bottomrule
% \end{tabular}
% \caption{OOD detector accuracy comparison on CIFAR-10, CIFAR-100, TinyImageNet and ImageNet datasets. All methods show a non-negligible amount of misclassification.}
% \label{tab:comparison_methods}
% \end{table}
\begin{table}[htbp]
\centering
\scriptsize
\renewcommand{\arraystretch}{0.8}
\setlength{\tabcolsep}{5pt}
\resizebox{\columnwidth}{!}{%
\begin{tabular}{lcccc}
\toprule
Method & CIFAR-10-C & CIFAR-100-C & TinyImageNet-C & ImageNet-C \\
\midrule
GMM classifier        & 74.73 & 70.74 & 59.23 & 55.42 \\
Entropy K-means       & 70.16 & 67.73 & 56.79 & 58.33 \\
Energy K-means        & 73.92 & 65.86 & 56.78 & 55.00 \\
Entropy (Best Thre.)  & 75.98 & 70.15 & 59.90 & 67.08 \\
Energy (Best Thre.)   & 75.43 & 68.79 & 60.54 & 65.42\\
\bottomrule
\end{tabular}%
}
\caption{OOD detector accuracy comparison on CIFAR-10, CIFAR-100, TinyImageNet and ImageNet datasets. All methods show a non-negligible amount of misclassification.}
\label{tab:comparison_methods}
\end{table}

\section{Mitigating the ID–OOD Tradeoff: The ROSETTA Framework}
We assume that the covariate shift in csID data is moderate, causing some degradation in classification accuracy while preserving the original class semantics. 
In this setting, csID samples remain classifiable, and test-time adaptation via entropy minimization can help maintain their performance under distribution shift. 
However, in the open-set TTA setting, this process introduces a trade-off between csID classification and csOOD detection. In particular, when csID samples are misclassified as csOOD, applying entropy maximization to these samples can severely degrade ID classification accuracy. 
We analyze this trade-off through two key observations and derive from them the design of ROSETTA, a framework that balances csID classification and csOOD detection.
\vspace{0.2cm}

\noindent\textbf{Entropy Minimization for csID Samples.} To better understand this trade-off, we first analyze the behavior of csID samples under entropy minimization.
Following prior test-time adaptation works such as TENT and UniEnt, entropy minimization is applied to samples predicted as csID within each test batch.
Let $\mathcal{B}_{\text{t,csID}}$ denote the detected csID samples, and $p_\theta(\mathbf{x})$ the softmax output of the model.
The loss is defined as
\begin{equation}
\begin{aligned}
\mathcal{L}_{\text{csID}}
&=
\frac{1}{\|\mathcal{B}_{\text{t,csID}}\|}
\sum_{\tilde{\mathbf{x}}_i \in \mathcal{B}_{\text{t,csID}}}
H(p_\theta(\tilde{\mathbf{x}}_i))
- \beta_1 H(\bar{p}_\theta), \\
&\text{where} \quad 
H(p) = -\sum_{k=1}^{K} p_k \log p_k.
\end{aligned}
\label{eq:l_csid}
\end{equation}

where $\bar{p}_\theta=\frac{1}{\|\mathcal{B}_t\|}\sum_{\tilde{\mathbf{x}}_i\in\mathcal{B}_t}p_\theta(\tilde{\mathbf{x}}_i)$.
The first term encourages confident predictions on csID samples, while the marginal-entropy term $H(\bar{p}_\theta)$ mitigates model collapse \cite{chen2022contrastive,choi2022improving,liang2020we,lim2023ttn}.

\newtheorem{observation}{Observation}
\begin{observation}\label{p1} Applying entropy minimization to pseudo-csID data in a model pre-trained on the ID dataset leads to an increase in the norm of the feature vectors.
\end{observation}
For a discriminative classifier, the logits for the target class satisfy $\psi_y(\mathbf{x}) > \psi_k(\mathbf{x})$ for all $k\neq y$. 
%When input $\tilde{\mathbf{x}}$ is perturbed with a bounded $\ell_p$-norm perturbation $\varepsilon \in \mathbb{R}_+$, the input $\tilde{\mathbf{x}}$ is located  outside an $\ell_p$ ball centered at $\mathbf{x}$, $B_p(\mathbf{x}, \varepsilon) := \{ \tilde{\mathbf{x}} \mid \|\tilde{\mathbf{x}} - \mathbf{x}\|_p \geq \varepsilon \}$. 
The largest logit of the input $\mathbf{x}$ is given by,
\begin{equation}
\psi_y(\mathbf{x}) = \max_y \|\mathbf{W}_y^{(L)}\| \|\mathbf{a}^{(l)}(\mathbf{x})\| \cos\theta_y.
\label{eq:logit}
\end{equation}
where $\cos\theta_y$ denotes the angle bewteen $\mathbf{W}_y^{(L)}$ and $\mathbf{a}^{(l)}(\mathbf{x})$. When covariate shift is introduced, the model's overall performance on the csID data degrades \cite{hendrycks2019benchmarking}. We use $\tilde{\mathbf{x}}$ to denote an input affected by covariate shift. Specifically, there exists some covariate-shifted input $\mathbf{\tilde{x}_0}$ such that
\begin{equation}
    \psi_y(\mathbf{x_0}) > \psi_y(\tilde{\mathbf{x}}_0).
\end{equation}

 The entropy of the model’s probability output for $\mathbf{\tilde{x}_0}$ is defined as,
\begin{equation}
    H(\mathcal{M}_\theta(\mathbf{\tilde{x}_0})) = - \sum_{k=1}^{K} \mathcal{M}_\theta^k(\tilde{\mathbf{x}}_0) \log \mathcal{M}_\theta^k(\tilde{\mathbf{x}}_0).
\end{equation}

Entropy minimization aims to restore prediction confidence by encouraging a more peaked output distribution. 
However, as Eq.~\ref{eq:logit} shows, since the classification weight $\mathbf{W}^{(L)}$ remains fixed during test-time adaptation, this confidence gain is primarily achieved by increasing the feature norm $\|\mathbf{a}^{(l)}(\mathbf{x})\|$, rather than through improved feature–prototype alignment characterized by a larger $\cos\theta_y$.
\vspace{0.2cm}

\begin{figure}[t]
    \centering
    \scriptsize
    \begin{subfigure}[b]{0.95\columnwidth}
        \centering
        \includegraphics[width=\linewidth]{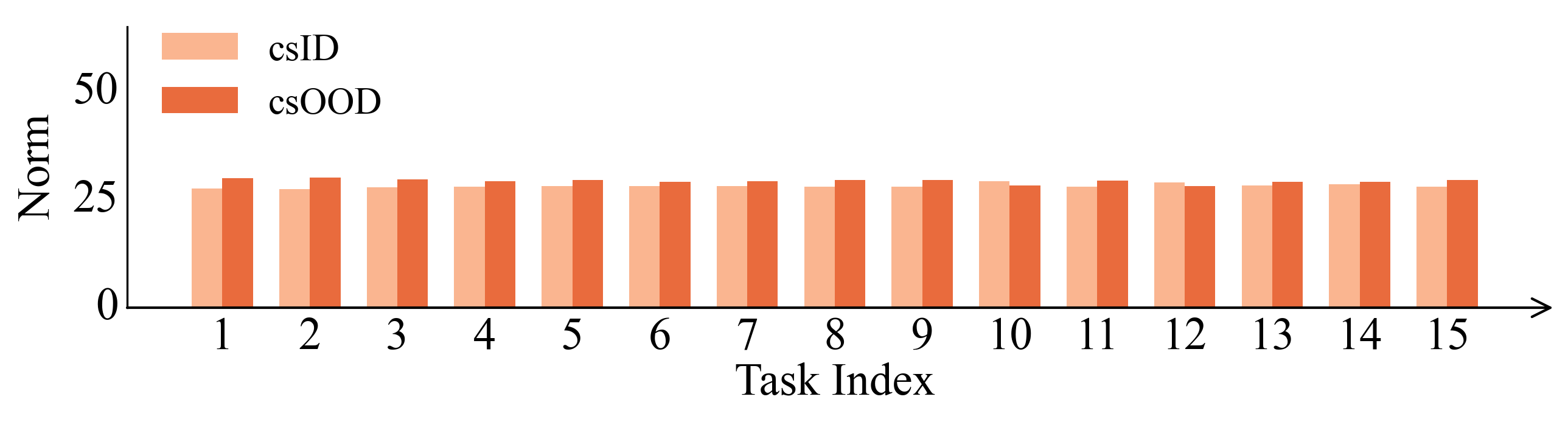}
        \caption{No Adaptation Applied.}
        \label{fig:figure2}
    \end{subfigure}

    \vspace{4pt}

    \begin{subfigure}[b]{0.95\columnwidth}
        \centering
        \includegraphics[width=\linewidth]{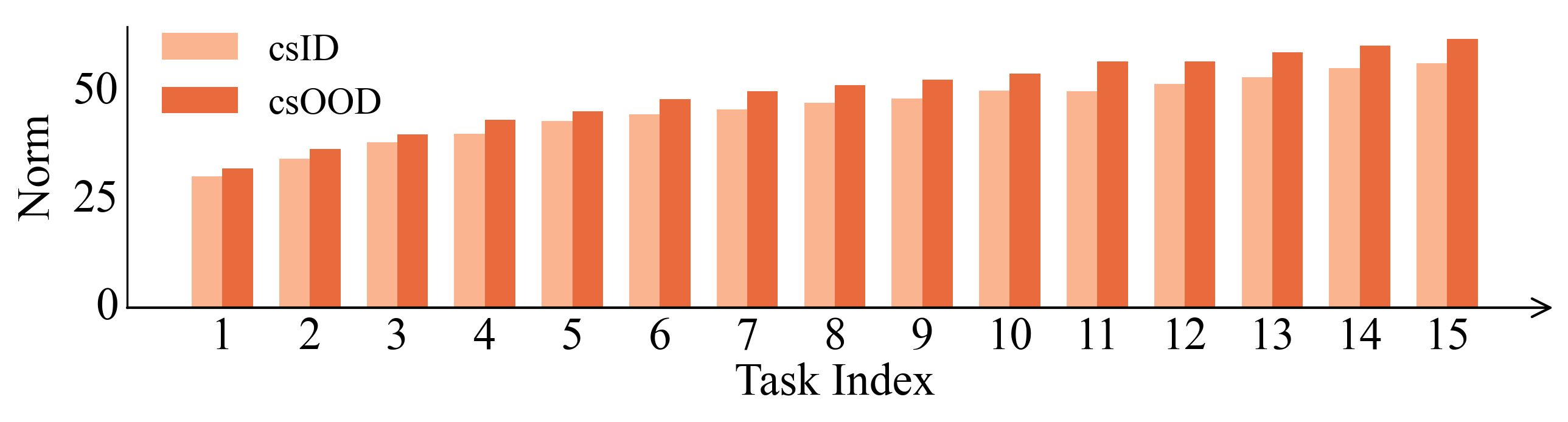}
        \caption{Adapt with $\mathcal{L}_\text{csID}$.}
        \label{fig:figure1}
    \end{subfigure}

    \vspace{4pt}

    \begin{subfigure}[b]{0.95\columnwidth}
        \centering
        \includegraphics[width=\linewidth]{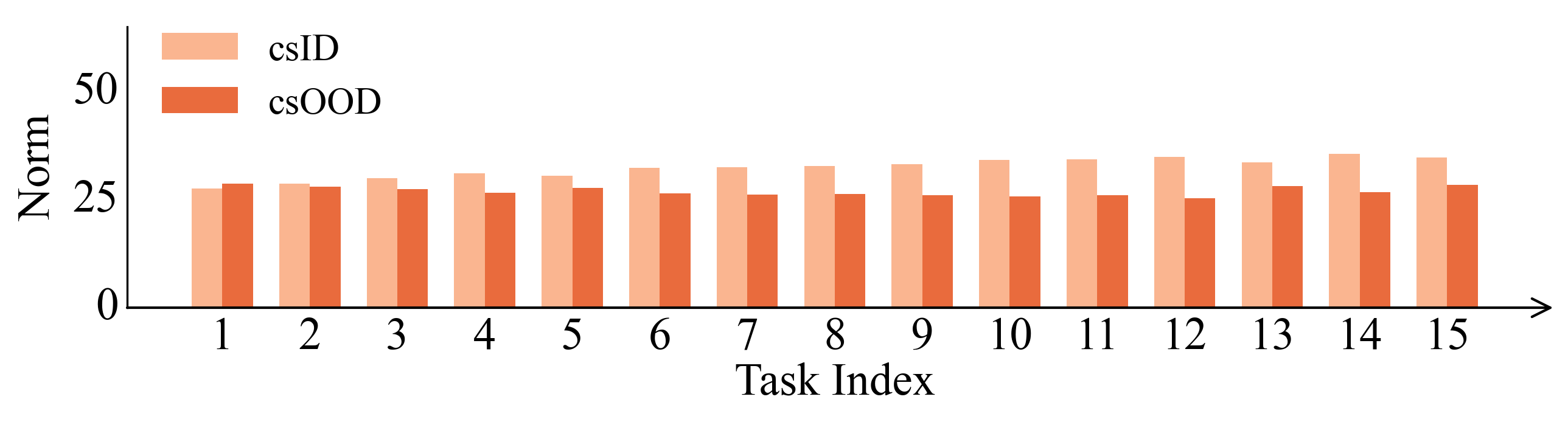}
        \caption{Adapt with $\mathcal{L}_\text{csID}+\mathcal{L}_\text{ang}$.}
        \label{fig:figure_cos}
    \end{subfigure}

    \caption{$l_2$-norm of feature vectors on CIFAR-10-C under various corruption types. The x-axis corresponds to corruption types, and the y-axis indicates the $l_2$-norm of features}
    \label{fig:three_images}
\end{figure}

\noindent\textbf{Empirical Verification.}
\, 
%Following UniEnt \cite{gao2024unient}, 
We empirically validate Observation~\ref{p1} using a WideResNet-40 model \cite{DBLP:conf/bmvc/ZagoruykoK16} pre-trained with AugMix \cite{hendrycks2020augmix} from RobustBench \cite{croce2021robustbench}.
 Experiments are conducted on CIFAR-10-C as the csID dataset and SVHN-C as the csOOD dataset. Entropy minimization is applied to pseudo-csID samples during adaptation. As shown in Fig.~\ref{fig:figure1}, the $l_2$-norm of features progressively increases under adaptation, while Fig.~\ref{fig:figure2} demonstrates that the feature norms remain stable in the absence of adaptation. These results verify that entropy minimization indeed increases the feature norms of all inputs, confirming our first observation. 
However, such uniform norm enlargement does not facilitate effective distinction between ID and OOD samples \cite{gao2024unient}.

\vspace{0.2cm}

\noindent\textbf{Motivation.}
While entropy minimization successfully enhances prediction confidence, its tendency to amplify feature norms may lead to overconfident representations for uncertain or shifted inputs.
To obtain stable and semantically consistent confidence, we aim to strengthen the directional alignment between csID features and their corresponding class prototypes instead of enlarging feature magnitude.
To this end, we introduce an \emph{angular loss} that explicitly increases $\cos\theta_y$ in Eq.~\ref{eq:logit}, allowing the model to boost classification confidence for csID samples without further enlarging the feature norm.

\vspace{0.2cm}

\noindent\textbf{Angular Loss for Increasing the Model’s Confidence in csID.} We maintain a running mean of the class prototype to accommodate distributional changes during adaptation.
Let $\mathcal{B}_\text{t,csID}^c$ denote the set of samples in this batch that belong to class $c$ among those identified as csID.
The class prototype $\mathbf{\mu}_{\text{t,c}}$ is given by:
\begin{equation}
\mathbf{\mu}_{\text{t,c}} = \frac{1}{\|\mathcal{B}_\text{t,csID}^c\|} 
\sum_{\tilde{\mathbf{x}}_i \in \mathcal{B}_\text{t,csID}^c} 
\mathbf{a}^{(l)}(\tilde{\mathbf{x}}_i).
\end{equation}
The running mean $\mathbf{\mu}_{\text{proto,c}}$ for each class $c$ is updated at each batch as:
\begin{equation}
\mathbf{\mu}_{\text{proto,c}} \leftarrow (1 - \alpha)\,\mathbf{\mu}_{\text{proto,c}} + \alpha\,\mathbf{\mu}_{\text{t,c}},
\end{equation}
where $\alpha$ is the momentum.
The angular loss is then defined as:
\begin{equation}
\mathcal{L}_{\text{ang}} = 
\frac{1}{\|\mathcal{B}_\text{t,csID}\|} 
\sum_{\tilde{\mathbf{x}}_i \in \mathcal{B}_{\text{t,csID}}}
\Big(1 - 
\frac{\mathbf{\mu}_{\text{proto},c} \cdot \mathbf{a}^{(l)}(\tilde{\mathbf{x}}_i)}
{\|\mathbf{\mu}_{\text{proto},c}\|\,\|\mathbf{a}^{(l)}(\tilde{\mathbf{x}}_i)\|}
\Big).
\label{eq:l_cos}
\end{equation}
Minimizing $\mathcal{L}_{\text{ang}}$ encourages csID features to align with their prototypes in direction, thereby enhancing classification confidence through angular consistency rather than norm escalation.
This stabilizes feature magnitudes during adaptation and lays the foundation for maintaining reliable csID/csOOD separation, as further analyzed in Observation~\ref{p2}. 
% \textcolor{red}{This completes our analysis on csID confidence. We next examine how feature norm relates to OOD separability, which forms the basis of the following observation.}
Having stabilized csID representations via angular alignment, we next examine how feature norm relates to OOD separability, which serves as the basis of the following observation.

\vspace{0.2cm}

\noindent\textbf{Feature Norm and OOD Detection.}
Prior studies have consistently observed a strong correlation between feature norm and OOD behavior 
\cite{dhamija2018reducing, yu2020out, chen2020norm, meng2021magface, park2023understanding}. 
Models trained on ID data tend to produce larger feature norms for ID samples compared to OOD samples, 
reflecting higher confidence in familiar regions of the feature space. 
This phenomenon has been leveraged in various contexts, such as face recognition and person search, 
to reject unseen or low-quality inputs based on their smaller activation magnitudes.
Motivated by these findings, we further analyze this property in the context of open-set TTA and formalize it in Observation~\ref{p2}.

\begin{observation}\label{p2}
To achieve effective OOD detection, the logits of csOOD samples should generally have a smaller $l_1$-norm compared to those of csID samples.
\end{observation}
Empirically, discriminative models often produce feature vectors with larger norms for ID samples and relatively smaller norms for OOD samples. This tendency can be explained by theoretical work showing that the feature norm serves as an implicit measure of confidence within the classifier’s hidden representation \cite{park2023understanding}. Specifically, the $l_1$-norm of a feature vector approximates the maximum logit of this implicit classifier. 
As a result, OOD samples exhibit smaller feature norms than ID samples, consistent with their lower confidence levels.
\vspace{0.2cm}

\noindent\textbf{Empirical Verification.}
We analyze the logits of csID and csOOD samples after entropy minimization adaptation.
For each sample, the logits are sorted in descending order, and the mean value at each rank position is computed across samples.
This statistical analysis captures the overall logit distribution for each group.
As shown in Fig.~\ref{fig:logits1}, csID samples exhibit a sharp decay in logits, forming near one-hot distributions, whereas csOOD samples display flatter curves, indicating less confident predictions.
% These results confirm that csOOD samples correspond to smaller logit magnitudes and consequently smaller feature norms.
\vspace{0.2cm}

\noindent\textbf{Motivation.}
Given that smaller feature norms are naturally associated with lower model confidence, explicitly enforcing this property for csOOD samples can enhance the model’s separability between csID and csOOD.
Building on this insight, we introduce a norm minimization strategy that suppresses the activation strength of features identified as csOOD, thereby improving OOD detection reliability.
\vspace{0.2cm}

\noindent\textbf{Logit Suppression for OOD Samples.}
To implement the above idea, we introduce an $l_1$-norm loss for samples identified as csOOD.
Let $\mathbf{a}^{(l)}$ denote the feature embedding at layer $l$. Its $l_1$-norm is defined as $\|\mathbf{a}^{(l)}\|_1 = \sum_{j=1}^{d_l} |a_j^{(l)}|$, where $d_l$ denotes the feature dimension.
We minimize this norm for csOOD samples through the following objective:
\begin{equation}
\mathcal{L}_{\text{norm}} = 
\frac{1}{\|\mathcal{B}_{\text{t,csOOD}}\|}
\sum_{\tilde{\mathbf{x}}_i \in \mathcal{B}_{\text{t,csOOD}}}
\|\mathbf{a}^{(l)}(\tilde{\mathbf{x}}_i)\|_1.
\end{equation}
Theoretical studies suggest that smaller feature norms for OOD samples improve detection performance \cite{park2023understanding}.
In our framework, this loss complements the angular loss introduced earlier. While $\mathcal{L}_{\text{ang}}$ enhances csID confidence through angular alignment, $\mathcal{L}_{\text{norm}}$ suppresses csOOD activations, ensuring a clear separation between the two groups.

\begin{figure}[t]
    \centering
    \includegraphics[width=0.45\textwidth]{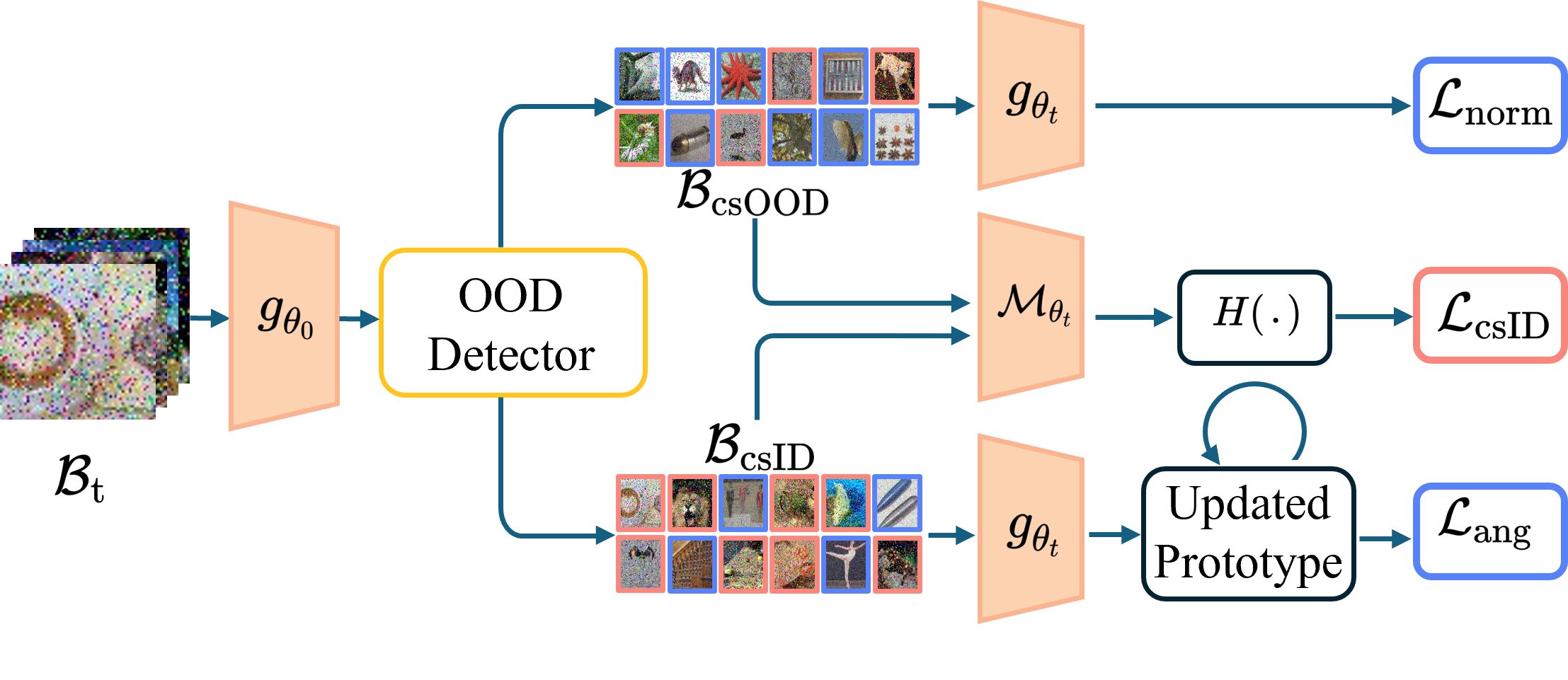}
    \caption{Overview of the ROSETTA pipeline.}
    \label{fig:pipeline}
\end{figure}

\begin{figure}[t]
    \centering
    \scriptsize
    \begin{subfigure}{0.48\linewidth}
        \centering
        \includegraphics[width=\linewidth]{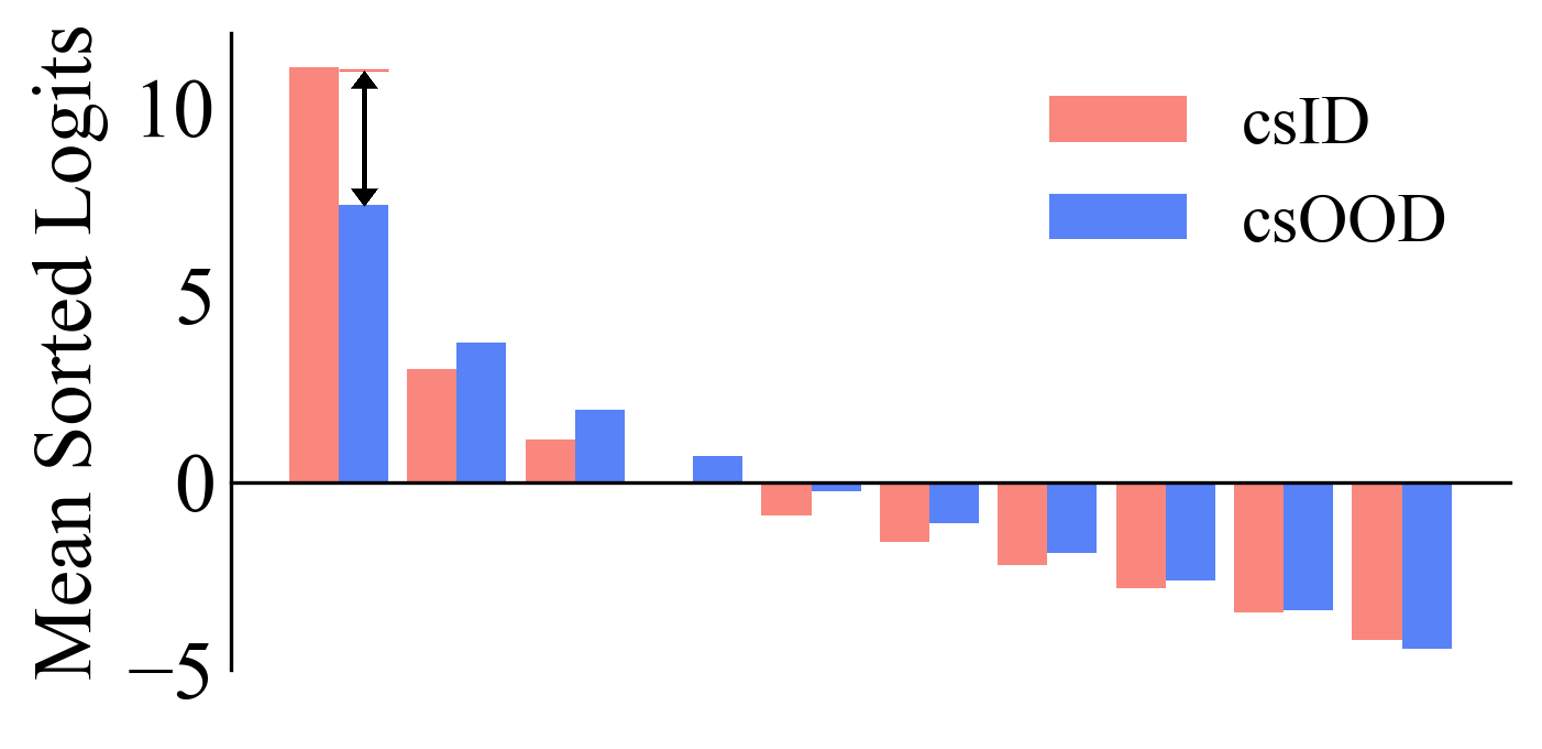}
        \caption{csID and csOOD logits w/ $\mathcal{L}_\text{csID}$.}
        \label{fig:logits1}
    \end{subfigure}
    \hfill
    \begin{subfigure}{0.48\linewidth}
        \centering
        \includegraphics[width=\linewidth]{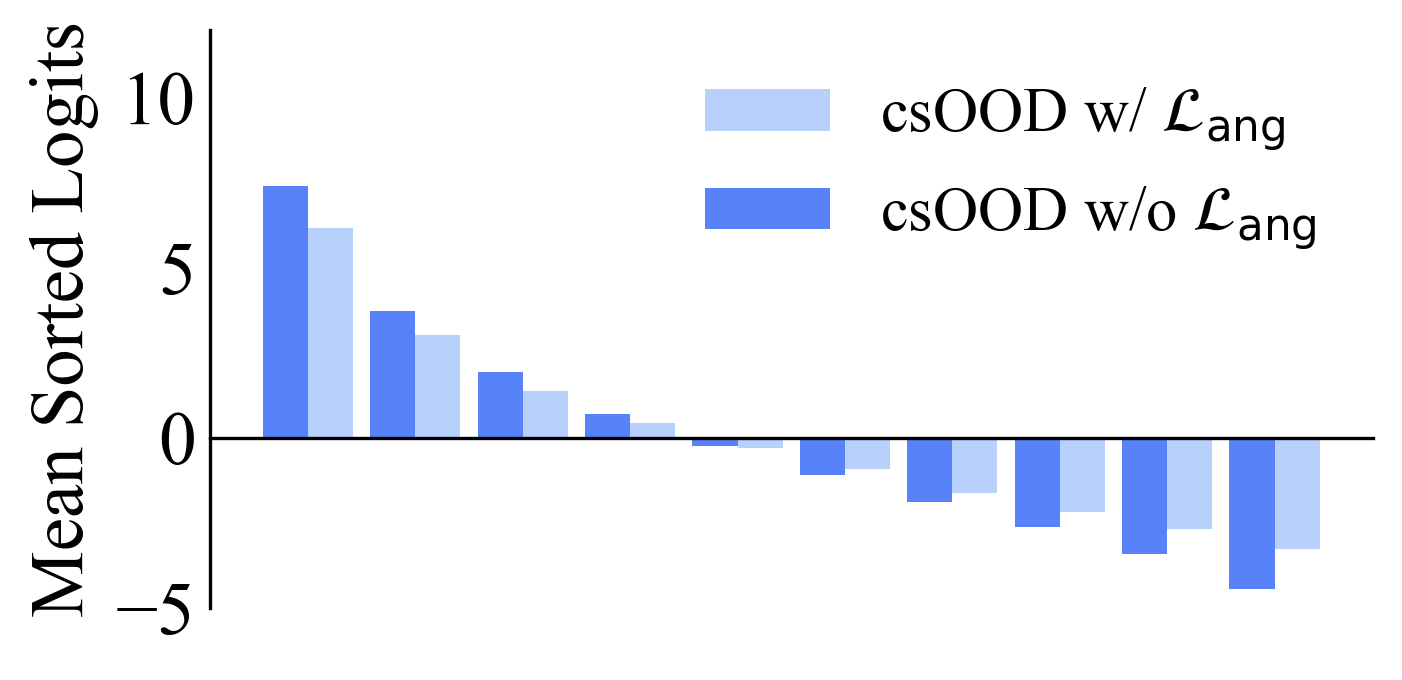}
        \caption{csOOD logits w/ and w/o $\mathcal{L}_\text{ang}$.}
        \label{fig:logits2}
    \end{subfigure}

    \vspace{6pt}

    \begin{subfigure}{0.48\linewidth}
        \centering
        \includegraphics[width=\linewidth]{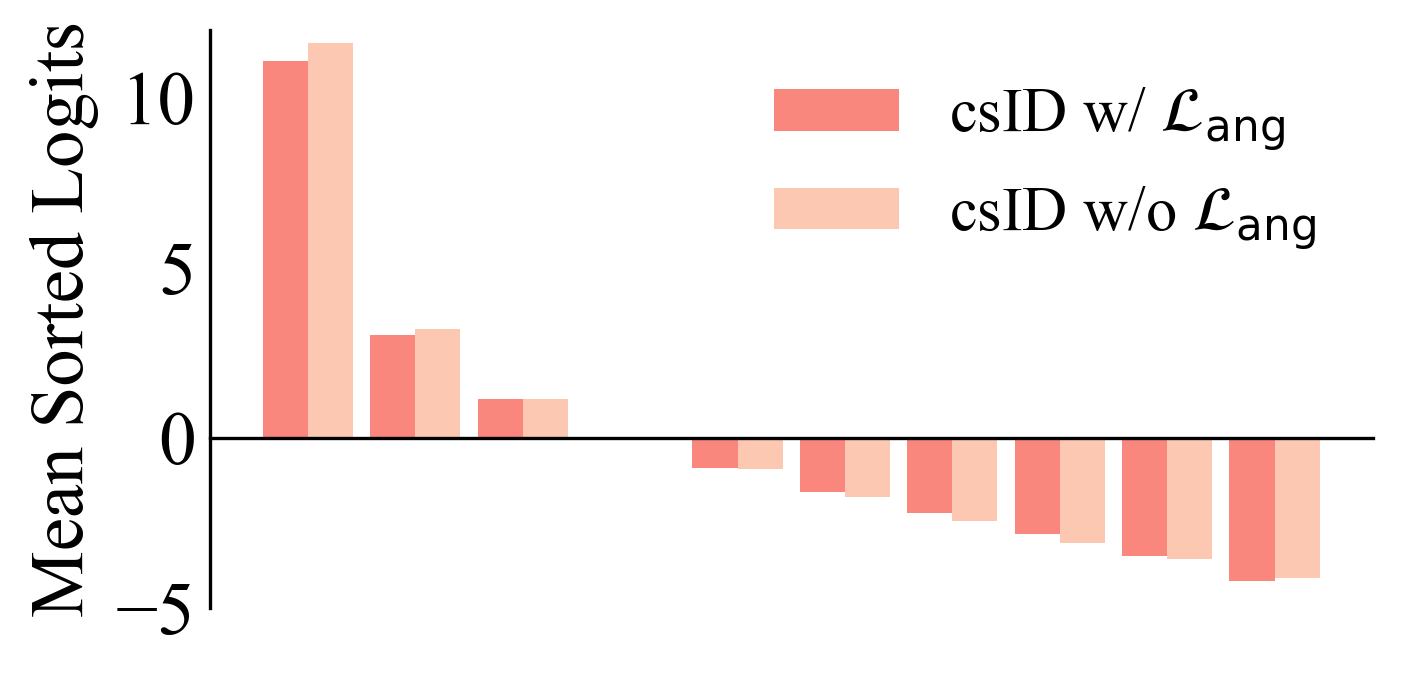}
        \caption{csID logits w/ and w/o $\mathcal{L}_\text{ang}$.}
        \label{fig:logits3}
    \end{subfigure}
    \hfill
    \begin{subfigure}{0.48\linewidth}
        \centering
        \includegraphics[width=\linewidth]{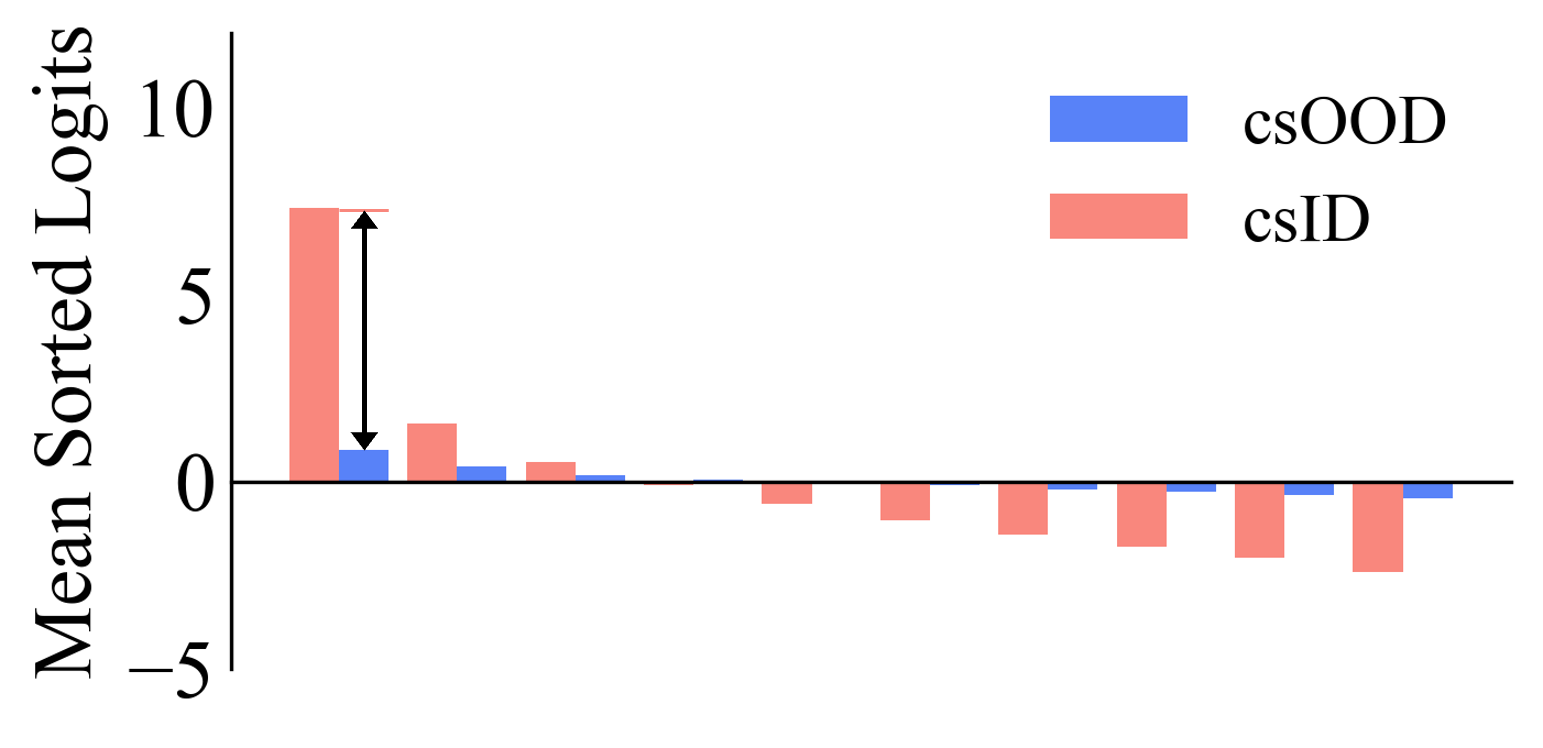}
        \caption{Logits adapted w/ $\mathcal{L}_\text{ROSETTA}$.}
        \label{fig:logits4}
    \end{subfigure}

    \caption{Mean sorted logits under different adaptation methods.}
    \label{fig:logits_combined}
\end{figure}

\vspace{4pt}
\noindent\textbf{Overall Objective.}
Together, the two complementary objectives form the core of our proposed method, \textbf{ROSETTA}.
We replace the conflicting entropy maximization term in open-set TTA with $\mathcal{L}_{\text{ang}}$ and $\mathcal{L}_{\text{norm}}$, thereby decoupling OOD detection from csID adaptation.
The overall loss is defined as:
\begin{equation}
\mathcal{L}_{\text{ROSETTA}} =
\mathcal{L}_{\text{csID}} + \gamma_1 \mathcal{L}_{\text{ang}} + \gamma_2 \mathcal{L}_{\text{norm}},
\end{equation}
where $\gamma_1$ and $\gamma_2$ are hyperparameters.
Following prior work \cite{wang2020tent}, we update only the batch normalization parameters during test-time adaptation for efficiency.
As illustrated in Fig.~\ref{fig:pipeline}, ROSETTA balances csID classification confidence and csOOD detection ability, effectively mitigating the trade-off discussed above.

% % 第二张图：四图组合

% The $l_1$-norm of \(\mathbf{a}^{(l)}\) is defined as \(\|\mathbf{a}^{(l)}\|_1 = \sum_{j=1}^{d_l} \left| a_j^{(l)} \right|\), where \(d_l\) denotes the feature dimension and \(a_j^{(l)}\) is the \(j^{\text{th}}\) element of \(\mathbf{a}^{(l)}\).

% Theoretical studies on feature norms for OOD detection suggest that a smaller $l_1$-norm for OOD samples leads to improved detection performance \cite{park2023understanding}. Moreover, the proposed loss is compatible with both the entropy and angular losses, which contributes to improved overall robustness.

% \noindent\textbf{ROSETTA.}
% Building on the two losses described above, we introduce the ROSETTA loss (Fig. \ref{fig:pipeline}), which preserves classification performance on csID samples while significantly improving performance in detecting csOOD samples. We replace the conflicting entropy maximization term with $\mathcal{L}_{\text{ang}}$ and $\mathcal{L}_{\text{norm}}$, thereby decoupling OOD detection from adaptation in the open-set TTA setting. The ROSETTA loss is defined as: 
% \begin{equation}
%     \mathcal{L}_{\text{ROSETTA}} = \mathcal{L}_{\text{csID}} +\gamma_1\mathcal{L}_{\text{ang}} +
%     \gamma_2\mathcal{L}_{\text{norm}}.
% \end{equation}
% where $\gamma_1$ and $\gamma_2$ are hyperparameters. In our ablation study, we show that ROSETTA is robust to the choice of these hyperparameters. Following prior works \cite{wang2020tent}, we update only the batch normalization parameters during test-time adaptation for parameter efficiency.

\section{Experiment} 

% \begin{table*}[!ht]

%The objective of these experiments is to demonstrate that our proposed DisCoNorm enhances csOOD detection capabilities while preserving csID classification performance. To this end, we evaluate DisCoNorm’s effectiveness across diverse datasets and in conjunction with other test-time adaptation methods. Additionally, we conduct an ablation study to assess the contribution of each component in the proposed loss function.
\subsection{Setup}
\noindent\textbf{Datasets.} Building on prior research, we assess our proposed method using well-established corruption benchmark datasets: CIFAR-10-C, CIFAR-100-C, Tiny-ImageNet-C and ImageNet-C \cite{hendrycks2019benchmarking}. Each dataset includes 15 types of corruption applied at 5 levels of severity. We focus on the highest level of corruption severity. Pre-trained models are trained on clean training sets, with testing and adaptation conducted on the corrupted test sets. To align with OSTTA \cite{lee2023towards}, we apply the same corruption types to the original SVHN \cite{netzer2011reading} and ImageNet-O \cite{hendrycks2021natural} test sets, producing the SVHN-C and ImageNet-O-C datasets. These datasets serve as covariate-shifted OOD sets for CIFAR-10/100-C and Tiny-ImageNet-C/ImageNet-C, respectively.

\noindent\textbf{Evaluation Protocols.}
% Following recent studies \cite{lee2023towards, niu2022efficient, wang2020tent, wang2022continual}, we evaluate TTA methods in continuously changing domains without resetting parameters between domains.
During testing, the corrupted images are fed to the model sequentially, and after each mini-batch, the parameters of the batch normalization layer are adapted. 
%Predictions at a given timestamp are unaffected by future data. 
For evaluation, we report accuracy (Acc) on csID data and evaluate csOOD detection using AUROC and the false positive rate at 95\% true positive rate (FPR95). 
% To demonstrate the model's overall performance on both csID and csOOD data, we also report the openset classification rate (OSCR) \cite{dhamija2018reducing} and H-score \cite{dong2025towards}.
To assess the model’s overall performance under the open-set setting, we adopt OSCR~\cite{dhamija2018reducing} and H-score \cite{dong2025towards}, two established metrics that provide a comprehensive evaluation of both classification and detection performance.

\noindent\textbf{Baseline Methods.}
We mainly compare our method with UniEnt and UniEnt+, the state-of-the-art methods for addressing open-set TTA. We also compare against other baselines, focusing on two types of prior TTA methods: 
\begin{itemize}
    \item \textbf{Entropy-free methods:} The source method evaluates test data using the source model without adaptation. BN Adapt \cite{nado2020evaluating} updates batch normalization statistics with test data during TTA. 
    % CoTTA adopts a teacher-student framework with stochastic restoration to reduce errors and forgetting.
   CoTTA employs a teacher-student framework to generate weight-averaged and augmentation-averaged pseudo-labels, reducing error accumulation, and applies stochastic restoration to counteract catastrophic forgetting. 
    \item \textbf{Entropy-based methods:} 
    TENT adapts by updating batch normalization parameters via entropy minimization. EATA selects reliable and diverse samples with low entropy and distinct outputs, and incorporates Fisher regularizer to prevent forgetting. 
    OSTTA filters out noisy data by using the wisdom of crowds to improve performance.
\end{itemize} 

% Our method can be integrated with existing entropy-based methods without further modifications. \textcolor{red}{ When applying it to EATA, we use our filtering approach, keeping all other components unchanged.}

\noindent\textbf{Implementation Details.}
We build on UniEnt (MIT License), adopting its experimental setups. Baseline results are obtained following UniEnt. Experiments were run on a 4×NVIDIA RTX A5000 machine with 512 GB RAM, each completing within 4 hours. Results are consistent across random seeds.
For CIFAR benchmarks, we use a WideResNet-40 pretrained with AugMix from RobustBench. For Tiny-ImageNet-C, we employ a ResNet-50 pretrained on Tiny-ImageNet and initialized with ImageNet weights, trained for 50 epochs using SGD (batch size 256, initial LR 0.01, cosine annealing).
For ImageNet-C, we use an AugMix-pretrained ResNet-50 from RobustBench. TTA uses Adam (LR 0.001 for TENT, EATA, OSTTA, 0.01 for CoTTA), with batch sizes of 200 (CIFAR, Tiny-ImageNet-C) and 64 (ImageNet-C).
Following T3A \cite{iwasawa2021test}, we treat the linear classifier weights as source prototypes, updating them via the mean embedding of csID samples in each mini-batch with momentum $\alpha = 0.005$.
The baseline results were obtained using the official UniEnt code. 
% In our experiments, the loss weights $\lambda_2$ and $\lambda_3$ are selected based on performance observed on the validation set.

\begin{table*}[!htbp]
    \scriptsize
    \centering
    % \caption{Performance comparison on CIFAR-10-C and CIFAR-100-C}
    % \begin{tabular*}{\linewidth}{@{\extracolsep{\fill}}r|cccc|cccc|cccc|cccc@{}}
    \renewcommand{\arraystretch}{0.8}  
    \resizebox{\textwidth}{!}{%
        \begin{tabular}{@{}r|cccc|cccc|cccc|cccc@{}}  % 使用 tabular 而非 tabular*
        \toprule
        \multirow{2}{*}{Method} & \multicolumn{4}{c}{CIFAR-10-C} & \multicolumn{4}{c}{CIFAR-100-C} & \multicolumn{4}{c}{Tiny-ImageNet-C} & \multicolumn{4}{c}{ImageNet-C} \\
        \cmidrule(lr){2-5} \cmidrule(lr){6-9} \cmidrule(lr){10-13} \cmidrule(lr){14-17}
        & \multicolumn{1}{c}{Acc$\uparrow$} & \multicolumn{1}{c}{AUROC$\uparrow$} & \multicolumn{1}{c}{FPR95$\downarrow$} & \multicolumn{1}{c}{OSCR$\uparrow$} 
        & \multicolumn{1}{c}{Acc$\uparrow$} & \multicolumn{1}{c}{AUROC$\uparrow$} & \multicolumn{1}{c}{FPR95$\downarrow$} & \multicolumn{1}{c}{OSCR$\uparrow$} 
        & \multicolumn{1}{c}{Acc$\uparrow$} & \multicolumn{1}{c}{AUROC$\uparrow$} & \multicolumn{1}{c}{FPR95$\downarrow$} & \multicolumn{1}{c}{OSCR$\uparrow$} 
        & \multicolumn{1}{c}{Acc$\uparrow$} & \multicolumn{1}{c}{AUROC$\uparrow$} & \multicolumn{1}{c}{FPR95$\downarrow$} & \multicolumn{1}{c}{OSCR$\uparrow$} \\
        \midrule
        Source & 81.73 & 77.89 & 79.45 & 68.44 & 53.25 & 60.55 & 94.98 & 39.87   & 22.29 & 53.79 & 93.41 & 16.29 & 28.21 & 49.63 & 94.74 & 19.81 \\
        BN Adapt & 84.20 & 80.40 & 76.84 & 72.13 & 57.16 & 72.45 & 84.29 & 47.10 & 37.00 & 61.06 & 90.90 & 28.50 & 43.57 & 55.89 & 93.39 & 30.42\\
        CoTTA & 85.77 & 85.89 & 72.40 & 77.26 & 56.46 & 77.04 & 80.96 & 48.95 & 12.68 & 41.36 & 94.43 & 7.61 & 47.67 & 55.58 & 94.51 & 33.8\\
        \midrule
        TENT & 79.38 & 65.39 & 95.94 & 56.73 & 54.74 & 65.00 & 94.79 & 42.24 & 28.96 & 49.78 & 95.96 & 19.02 & 45.82 & 51.34 & 96.47 & 30.33\\
        + UniEnt & 84.31 & 92.28 & 36.74 & 80.32 & 59.07 & 89.28 & 51.14 & 56.26 & 37.23 & 63.92 & 89.72 & \textbf{30.18} & 47.53 & 56.33 & 95.21 & 34.42 \\
        + UniEnt+ & 84.03 & 93.18 & 32.74 & 80.62 & 58.58 & 91.39 & 41.09 & 56.36 & 37.31 & 63.83 & 89.12 & 30.12 & 46.87  & 55.86  & 95.10 & 33.73\\
        \rowcolor{gray!20}
        + Ours & \textbf{84.34} & \textbf{93.75} & \textbf{29.91} & \textbf{81.37} & 
        % \textbf{59.20} & \textbf{91.80} & \textbf{35.89} & \textbf{56.80}
        \textbf{59.39} & \textbf{91.54} & \textbf{37.30} & \textbf{56.86}
        & \textbf{37.32} & \textbf{64.01} & \textbf{88.53} & 29.75  & \textbf{48.01} & \textbf{57.35} & \textbf{95.08}  & \textbf{35.01}  \\
        \midrule
        EATA & 80.92 & 84.32 & 71.66 & 72.63 & 60.63 & 88.64 & 50.18 & 57.24 & 37.09 & 57.55 & 93.22 & 27.91 & 51.40 & 53.10 & 95.18 & 34.87\\
        + UniEnt & 84.31 & 97.15 & 13.25 & 82.99 & 59.75 & 93.42 & 30.36 & 57.99 & 37.54 & 64.34 & \textbf{89.23} & 30.59 & 49.60 & 58.29 & 93.63 & 36.28\\
        + UniEnt+ & 85.18 & 96.97 & 14.28 & 83.67 & 59.71 & 94.23 & 26.87 & 58.19 & 38.65 & 62.30 & 90.88 & 30.95 & \textbf{51.57} & 59.45 & 93.60 & \textbf{38.27}\\
        \rowcolor{gray!20}
        + Ours & \textbf{85.42} & \textbf{97.53} & \textbf{11.94} & \textbf{84.25}
                & \textbf{61.76} & \textbf{94.61} & \textbf{25.04} & \textbf{60.21}
                & \textbf{38.96} & \textbf{64.49} & 89.56 & \textbf{31.46} & 50.65 & \textbf{60.37} & \textbf{93.50} & 38.16\\
        \midrule
        OSTTA & 84.44 & 72.74 & 77.02 & 65.17 & 60.03 & 75.37 & 82.75 & 51.35 & 37.29 & 55.66 & 94.34 & 27.74 & 47.91 & 52.93 & 96.15 & 32.77\\
        + UniEnt & 82.46 & 96.20 & 16.37 & 80.51 & 58.69 & 94.84 & 22.95 & 57.28 & 33.72 & 62.29 & 89.67 & 26.63 & 47.92 & 56.02 & 95.23 & 34.47\\
        + UniEnt+ & 84.30 & \textbf{97.38} & \textbf{11.56} & 82.91 & 58.93 & \textbf{95.42} & \textbf{20.59} & 57.69 & 34.47 & 61.28 & \textbf{89.56} & 26.65 & 47.47 & 55.67 & 95.16 & 34.03\\
        \rowcolor{gray!20}
        + Ours & \textbf{85.56} & 96.12 & 18.96 & \textbf{83.46} & \textbf{60.06} & 93.06 & 31.44 & \textbf{58.02} & \textbf{38.85} & \textbf{62.30} & 91.08 & \textbf{30.73} & \textbf{47.95} & \textbf{57.70} & \textbf{94.47} & \textbf{34.91}\\
        \bottomrule
    \end{tabular}
    % \end{tabular*}
    }
    \caption{Results of the baseline methods and the proposed ROSETTA (Ours) on CIFAR-C benchmarks and ImageNet-C benchmarks. FPR95 indicates the False Positive Rate when the True Positive Rate is 95\%. \emph{The results demonstrate that our method achieves strong performance in both csID classification and csOOD detection simultaneously.}}
    \vspace{-8pt}
\label{table1_compareSOTA}
\end{table*}

\newcommand{\cmark}{\textcolor{green}{\ding{51}}} % Green check mark
\newcommand{\xmark}{\textcolor{red}{\ding{55}}}   % Red cross mark
\begin{table*}[!ht]
    \scriptsize
    \centering
    \renewcommand{\arraystretch}{0.8}  
    \begin{tabular*}{\linewidth}{@{\extracolsep{\fill}}r|ccc|cccc|cccc|}
        \toprule
        \multirow{2}{*}{Based Method} & \multirow{2}{*}{$\mathcal{L}_{\text{csID}}$} & \multirow{2}{*}{$\mathcal{L}_{\text{ang}}$} & \multirow{2}{*}{$\mathcal{L}_{\text{norm}}$} & \multicolumn{4}{c|}{CIFAR-10-C} & \multicolumn{4}{c|}{CIFAR-100-C} \\ 
        \cmidrule(lr){5-8} \cmidrule(lr){9-12}
        % \cmidrule(lr){5-12}
        &  &  & & Acc$\uparrow$& AUROC$\uparrow$ & FPR95$\downarrow$ & OSCR$\uparrow$ & Acc$\uparrow$ & AUROC$\uparrow$ & FPR95$\downarrow$ & OSCR$\uparrow$ \\
        \midrule
        \multirow{4}{*}{TENT+Ours} & \xmark & \xmark  & \xmark & 79.38 & \ 65.39 & 95.94 & 56.73 & 54.74 & 65.00 & 94.79 & 42.24\\
        & \cmark & \xmark & \xmark  & 85.04 & 81.80  & 68.89 & 73.57 & 59.30  & 86.09 & 63.65 & 55.55 \\
        & \cmark & \cmark & \xmark    & \textbf{85.66} & 85.63  & 65.67 & 77.32 & \textbf{60.88}  & 85.97 & 65.05 & 56.90 \\
        & \cmark & \cmark & \cmark    & 84.34 & \textbf{93.75} & \textbf{29.91} & \textbf{81.37} & 59.20 & \textbf{91.80} & \textbf{35.89} & \textbf{56.80} \\
        \midrule
        \multirow{4}{*}{EATA+Ours} & \xmark & \xmark  & \xmark  & 80.92 & 84.32  & 71.66 & 72.63 & 60.63  & 88.64 & 50.18 & 57.24 \\
        & \cmark & \xmark & \xmark   & 85.53 & 82.94  & 67.95 & 74.85 & 60.46  & 88.53 & 54.30 & 57.26 \\
        & \cmark & \cmark & \xmark    & \textbf{86.54} & 87.79  & 60.73 & 79.48 & \textbf{62.42} & 88.05 & 58.64 & 58.92 \\
        & \cmark & \cmark & \cmark    & 85.42 & \textbf{97.53} & \textbf{11.94} & \textbf{84.25}& 61.76 & \textbf{94.61} & \textbf{25.04} & \textbf{60.21}\\
        \bottomrule
    \end{tabular*}
    \caption{Comparison of methods on CIFAR-10-C and CIFAR-100-C datasets w/ and w/o $\mathcal{L}_{\text{csID}}$, $\mathcal{L}_{\text{ang}}$, and the $l_1$-norm loss $\mathcal{L}_{\text{norm}}$.}
    \vspace{-6pt}
    \label{table:ablation}
\end{table*}

\subsection{Main Results}
\noindent\textbf{CIFAR-C Benchmarks.}
We evaluate the ROSETTA loss on CIFAR-C benchmarks, with results presented in Table \ref{table1_compareSOTA}. A trade-off exists in state-of-the-art methods such as UniEnt and UniEnt+, where one method may perform better on ID classification while the other excels in OOD detection. However, neither method achieves the best performance on both metrics simultaneously. 
%This is caused by the inherent limitations of their methods. 
In contrast, our method consistently attains the highest performance on both ID classification accuracy and OOD detection (AUROC). For example, for TENT on CIFAR-100-C, UniEnt+ surpasses UniEnt for 2.11\% on AUROC, while UniEnt surpasses UniEnt+ for 0.49\% on ID accuracy. For our method, we can achieve the best performance in terms of both metrics.

Moreover, our approach improves both UniEnt and UniEnt+ on CIFAR-C benchmarks. Specifically, we observe OSCR improvements of 0.75\% and 0.5\% for TENT on CIFAR-10-C and CIFAR-100-C, respectively, and 0.58\% and 2.02\% for EATA on CIFAR-10-C and CIFAR-100-C.

\noindent\textbf{Tiny-ImageNet-C \& ImageNet-C Benchmarks.}
We further evaluate ROSETTA on the more challenging Tiny-ImageNet-C, with results shown in Table \ref{table1_compareSOTA}. Our method achieves performance comparable to UniEnt and UniEnt+ in terms of ID accuracy.
Notably, while UniEnt+ surpasses UniEnt in ID accuracy by 1.11\%, UniEnt outperforms UniEnt+ by 2.04\% in AUROC, suggesting that each method excels in a single metric. 
In contrast, our approach demonstrates balanced, competitive results across both ID and OOD metrics, matching UniEnt+ in ID accuracy and UniEnt in OOD detection performance. 

Furthermore, to evaluate the performance of our method on large-scale datasets, we conduct experiments on ImageNet-C. To construct the csOOD dataset, ImageNet-O-C, we apply common corruptions and perturbations to ImageNet-O, following the official implementation. We integrate our approach with TENT, EATA, and OSTTA, and the results are summarized in Table \ref{table1_compareSOTA}. As shown, ROSETTA achieves superior csOOD detection performance compared to UniEnt and UniEnt+, while maintaining comparable csID classification accuracy.
% To assess scalability, we conduct experiments on ImageNet-C and ImageNet-O-C. ROSETTA demonstrates superior csOOD detection and competitive csID accuracy compared to UniEnt and UniEnt+ when incorporated into the TENT, EATA, and OSTTA frameworks, as shown in Table~\ref{table1_compareSOTA}.

\begin{table}[t]
\centering
\scriptsize % 字体缩小一点（9pt）
\renewcommand{\arraystretch}{0.8}  
\setlength{\tabcolsep}{5pt} % 减小列间距
\begin{tabular}{l|cccc}
\toprule
Condition & Fog & Night & Rain & Snow\\
\midrule
TENT %\cite{wang2020tent}       
& 69.17 & 40.70 & 58.68 & 57.16 \\
+UniEnt %\cite{gao2024unient}   
& 69.13 \,(-0.04) & 40.60 \,(-0.10) & 58.68 \,(+0.00) & 57.02 \,(-0.14) \\
+UniEnt+ %\cite{gao2024unient}  
& 69.15 \,(-0.02) & 40.64 \,(-0.06) & 58.59 \,(-0.09) & 56.99 \,(-0.17) \\
\rowcolor{gray!20}
+Ours                       
& \textbf{69.24 \,(+0.07)} & \textbf{40.71 \,(+0.01)} & \textbf{58.92 \,(+0.24)} & \textbf{57.41 \,(+0.25)} \\
\bottomrule
\end{tabular}
\caption{Semantic segmentation results (mIoU) on the Cityscapes-to-ACDC test-time adaptation task.}
\label{semantic segmentation}
\end{table}

% \vspace{-6pt}
\subsection{Semantic Segmentation on Real-World Dataset}
% CIFAR-C and ImageNet-C are synthetic benchmarks designed to simulate noisy real-world scenarios.
To further demonstrate the effectiveness of the proposed method for real-world applications such as autonomous driving, we conduct an experiment on the semantic segmentation Cityscapes-to-ACDC task. In particular, the segmentation model Segformer-B5 \cite{xie2021segformer} is pre-trained on the Cityscapes dataset \cite{Cordts2016Cityscapes}, and adapted to the Adverse Conditions Dataset (ACDC) \cite{sakaridis2021acdc} which has four different adverse conditions: Fog, Night, Rain and Snow. In ACDC, we consider classified pixels as csID and void pixels as inherent csOOD. To better reflect practical deployments, we make a single pass over 400 unlabeled images for each type of noise. As shown in Table~\ref{semantic segmentation}, our method improves mIoU on the Rain and Snow conditions by 0.24\% and 0.25\% respectively. These gains illustrate the effectiveness of our approach for semantic segmentation in real-world settings. More implementation details can be found in Supplementary.

\begin{table*}[htbp]
\centering
\scriptsize
\setlength{\tabcolsep}{3pt}
\renewcommand{\arraystretch}{0.8}  
\begin{tabular}{l|ccc c|ccc c|ccc c|ccc c}
\toprule
 & \multicolumn{4}{c|}{H $\rightarrow$ A} & \multicolumn{4}{c|}{H $\rightarrow$ C} & \multicolumn{4}{c|}{C $\rightarrow$ A} & \multicolumn{4}{c}{C $\rightarrow$ H} \\
Method & Acc↑ & FPR95↓ & AUROC↑ & H-score↑ 
       & Acc↑ & FPR95↓ & AUROC↑ & H-score↑ 
       & Acc↑ & FPR95↓ & AUROC↑ & H-score↑ 
       & Acc↑ & FPR95↓ & AUROC↑ & H-score↑ \\
\midrule
Source     & 57.28 & 87.53 & 45.90 & 25.12 
           & 38.79 & 98.99 & 22.49 & 2.83  
           & 67.88 & 65.23 & 70.81 & 52.07 
           & 61.36 & 71.67 & 67.72 & 45.21 \\
Tent       & 57.51 & 78.59 & 65.63 & 37.82 
           & 41.64 & 94.03 & 47.25 & 14.11 
           & 66.78 & 80.68 & 63.56 & 36.38 
           & 62.94 & 78.73 & 65.22 & 38.35 \\
SAR        & 58.50 & 77.70 & 68.39 & 39.19 
           & 46.32 & 95.04 & 48.67 & 12.31 
           & 66.67 & 63.91 & 70.11 & 52.66 
           & 66.91 & 74.41 & 66.88 & 43.49 \\
OSTTA      & 55.63 & 76.82 & 72.14 & 40.01 
           & 44.49 & 89.98 & 58.36 & 21.52 
           & 65.56 & 73.73 & 64.03 & 43.52 
           & 63.52 & 78.30 & 64.44 & 38.79 \\
UniEnt     & 60.15 & 78.04 & 64.02 & 38.57 
           & 40.99 & 98.71 & 31.86 & 3.61  
           & 67.55 & 54.53 & 79.28 & 60.72
           & 66.26 & 64.38 & 76.11 & 53.28 \\
READ       & 57.84 & 68.87 & 69.58 & 47.03 
           & 45.40 & 86.95 & 56.73 & 25.80 
           & 63.69 & 68.65 & 68.72 & 48.27 
           & 61.07 & 69.79 & 69.55 & 46.98 \\
AEO        & 59.93 & 61.92 & 75.09 & 53.31 
           & 44.21 & 82.54 & 60.34 & 31.11 
           & 65.89 & 60.60 & 74.73 & 55.62 
           & 64.24 & 58.47 & 79.42 & 57.34 \\
Ours       & \cellcolor{gray!20} 59.93 &             \cellcolor{gray!20} 58.39 & \cellcolor{gray!20}78.16 & \cellcolor{gray!20}56.06 
           & \cellcolor{gray!20}44.49 & \cellcolor{gray!20}80.79 & \cellcolor{gray!20}63.44 & \cellcolor{gray!20}33.22
           & \cellcolor{gray!20}67.66 & \cellcolor{gray!20}57.06 & \cellcolor{gray!20}77.74 & \cellcolor{gray!20}58.90 
           & \cellcolor{gray!20}66.69 & \cellcolor{gray!20}47.58 & \cellcolor{gray!20}82.26 & \cellcolor{gray!20}64.90 \\
\bottomrule
\end{tabular}
\caption{Multimodal open-set TTA on the HAC dataset with video and audio modalities. Our method achieves strong performance across diverse domains.}
\label{tab:hac_subset}
\end{table*}

\subsection{Applicability Across Different Open-Set TTA Frameworks}
To demonstrate the general applicability of our method, we further apply it to AEO, a multimodal open-set TTA framework. We conduct experiments using the HAC \cite{dong2023simmmdg} dataset as the csID and EPIC-KITCHENS \cite{damen2018scaling} as the OOD. 
% The HAC dataset contains seven actions performed by humans (H), animals (A), and cartoon characters (C), forming three domains: H, A, and C. The adaptation setting H $\rightarrow$ A refers to adapting from the human domain to the animal domain.

The HAC dataset is a multimodal benchmark designed for cross-domain action recognition and adaptation. 
It contains seven action categories such as walking, jumping, and pushing, performed by three types of subjects: humans (H), animals (A), and cartoon characters (C). 
Each domain includes paired RGB videos and textual descriptions, enabling evaluation under multimodal and distribution-shifted conditions. 
Following the convention in \cite{dong2023simmmdg}, the adaptation setting H$\rightarrow$A refers to adapting from the human domain to the animal domain, where visual appearance and motion dynamics differ substantially.

For fair comparison, we reproduce the results of AEO reported in the table. Experimental results across various adaptation settings demonstrate consistent improvements in both csID classification and csOOD detection performance. These results suggest that our method can enhance a variety of open-set TTA approaches, confirming its robustness and broad applicability.

% \vspace{-6pt}
\subsection{Ablation Studies and Analysis}
\vspace{-0.1cm}
% We demonstrate the effectiveness of each component in $\mathcal{L}_{\text{ROSETTA}}$, analyze hyperparameter sensitivity, examine the trade-off between csID classification and csOOD detection, and present qualitative results using t-SNE in Supplementary.
% \textcolor{red}{Can we mention t-SNE at here?}
\vspace{0.1cm}

\noindent\textbf{Effectiveness of $\mathcal{L}_{\text{ang}}$.}
% Entropy minimization loss encourages the model to produce higher logits. However, this is often achieved by uniformly increasing the feature norm across all inputs, as shown in the comparison between Fig. \ref{fig:figure2} and Fig. \ref{fig:figure1}. While this boosts confidence on ID samples, increasing the norm of OOD samples also raises their maximum logits, which hinders effective OOD detection. 
Entropy minimization loss tends to increase feature norms uniformly, raising logits for both csID and csOOD samples, which hurts OOD detection (Fig. \ref{fig:figure1} vs Fig. \ref{fig:figure2}). $\mathcal{L}_{\text{ang}}$ encourages the model to achieve higher logits by increasing cosine similarity between csID features and class prototypes. Our approach reduces the feature norm of csOOD samples (Fig. \ref{fig:figure_cos}), resulting in a more uniform logit distribution for OOD samples (Fig. \ref{fig:logits2}), thereby improving OOD detection. Additionally, the increased cosine similarity with the updated class prototype improves csID classification accuracy. Table \ref{table:ablation} shows that incorporating $\mathcal{L}_{\text{ang}}$ enhances both csOOD detection and csID classification compared to using only $\mathcal{L}_{\text{csID}}$.

% \begin{figure}% 表示强制放置在当前位置
%     \centering
%     \includegraphics[width=0.8\linewidth]{image/ood_gradually.png}  % 替换为你图片的文件名
%     \caption{Impact of scaled OOD loss weight with $\tau$ on ID classification and OOD detection. As OOD loss weight increases, UniEnt and UniEnt+ show improved OOD detection but compromised ID classification, while our method maintains superior performance in both.}  % 图片标题
%     \label{fig:weight of ood loss}  % 设置标签，用于引用
% \end{figure}
% \vspace{0.1cm}

% \begin{figure}[t]
%     \centering
%     % 第一行
%     \begin{subfigure}[b]{0.495\linewidth}
%         \centering
%         \includegraphics[width=1\linewidth]{image/heatmap_cifar10.png}  % 替换为你的图片路径
%         % \caption{}
%         \vspace{-9pt}
%         \label{fig:heatmapa}
%     \end{subfigure}
%     \hfill
%     \begin{subfigure}[b]{0.495\linewidth}
%     \centering
%     \includegraphics[width=1\linewidth]{image/heatmap_cifar100.png}  % 替换为你的图片路径
%     % \caption{}
%     \vspace{-9pt}
%     \label{fig:heatmapb}
%     \end{subfigure}
%     \caption{Heatmap showing OSCR performance of ROSETTA with TENT adaptation across hyperparameter settings on CIFAR-10-C and CIFAR-100-C. 
%     % Color ranges from white to red to black, with higher OSCR values indicating better open-set recognition. 
%     % (a) CIFAR10-C results; (b) CIFAR100-C results, 
%     Demonstrating ROSETTA's robustness across different settings.}
%     % \vspace{-2pt}
%     \label{fig:heatmap}
% \end{figure}

\noindent\textbf{Effectiveness of $\mathcal{L}_{\text{norm}}$.}
The introduction of $\mathcal{L}_{\text{norm}}$ directly reduces the feature norm of csOOD samples, leading to a more uniform logits distribution for both csOOD and csID samples, which further enhances csOOD detection. As shown in Fig. \ref{fig:logits4}, $\mathcal{L}_{\text{norm}}$ promotes a more uniform distribution for csOOD. While the maximum logits of csID samples also decrease slightly, the gap between the maximum logits of csID and csOOD is significantly widened, greatly benefiting csOOD detection. Table \ref{table:ablation} demonstrates that incorporating $\mathcal{L}_{\text{norm}}$ substantially improves csOOD detection performance.

\begin{figure}[t]
    \centering
    \begin{subfigure}[b]{0.49\linewidth}
        \centering
        \includegraphics[width=\linewidth]{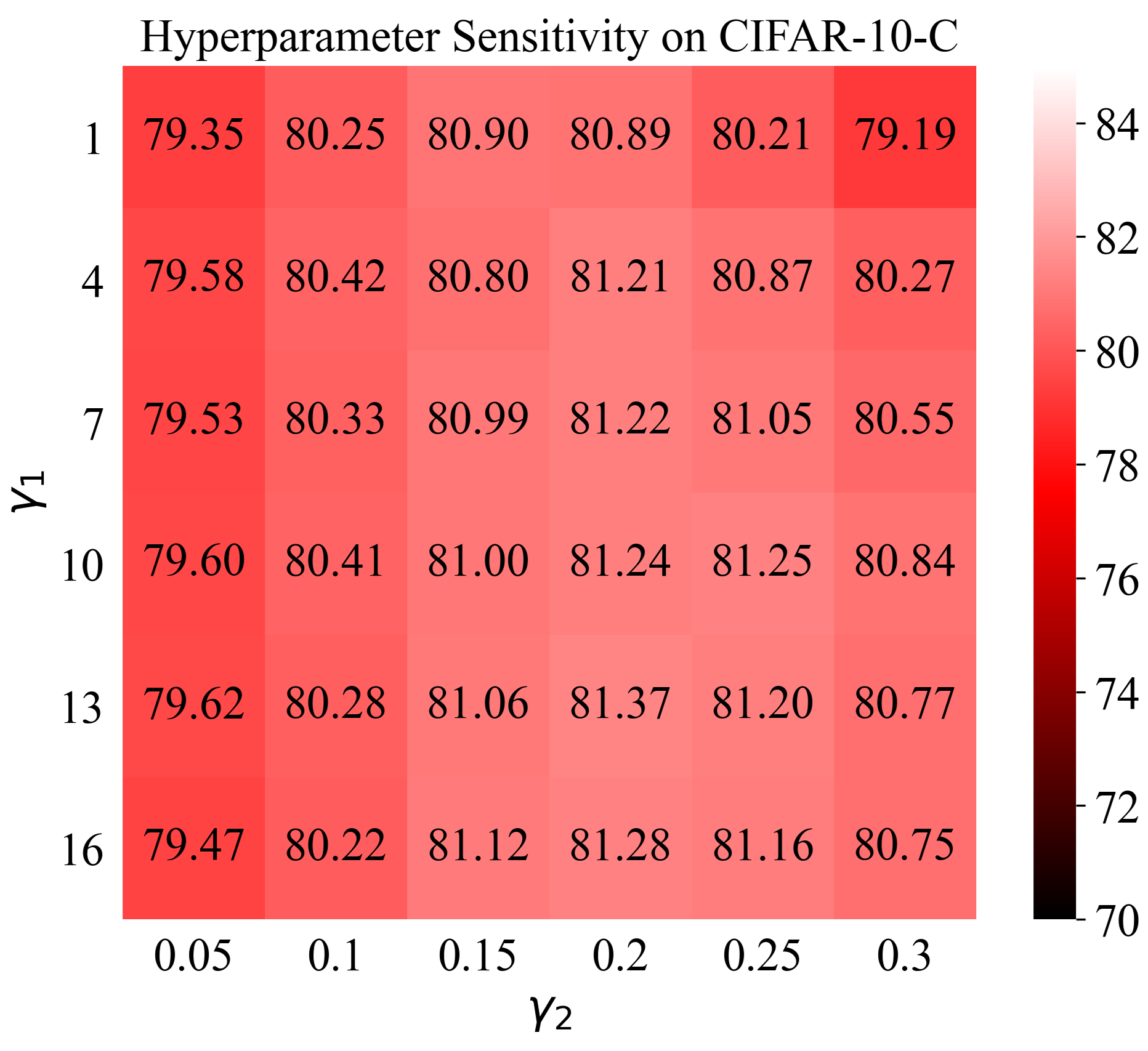}
        \vspace{-8pt}
        \label{fig:heatmapa}
    \end{subfigure}
    \hfill
    \begin{subfigure}[b]{0.49\linewidth}
        \centering
        \includegraphics[width=\linewidth]{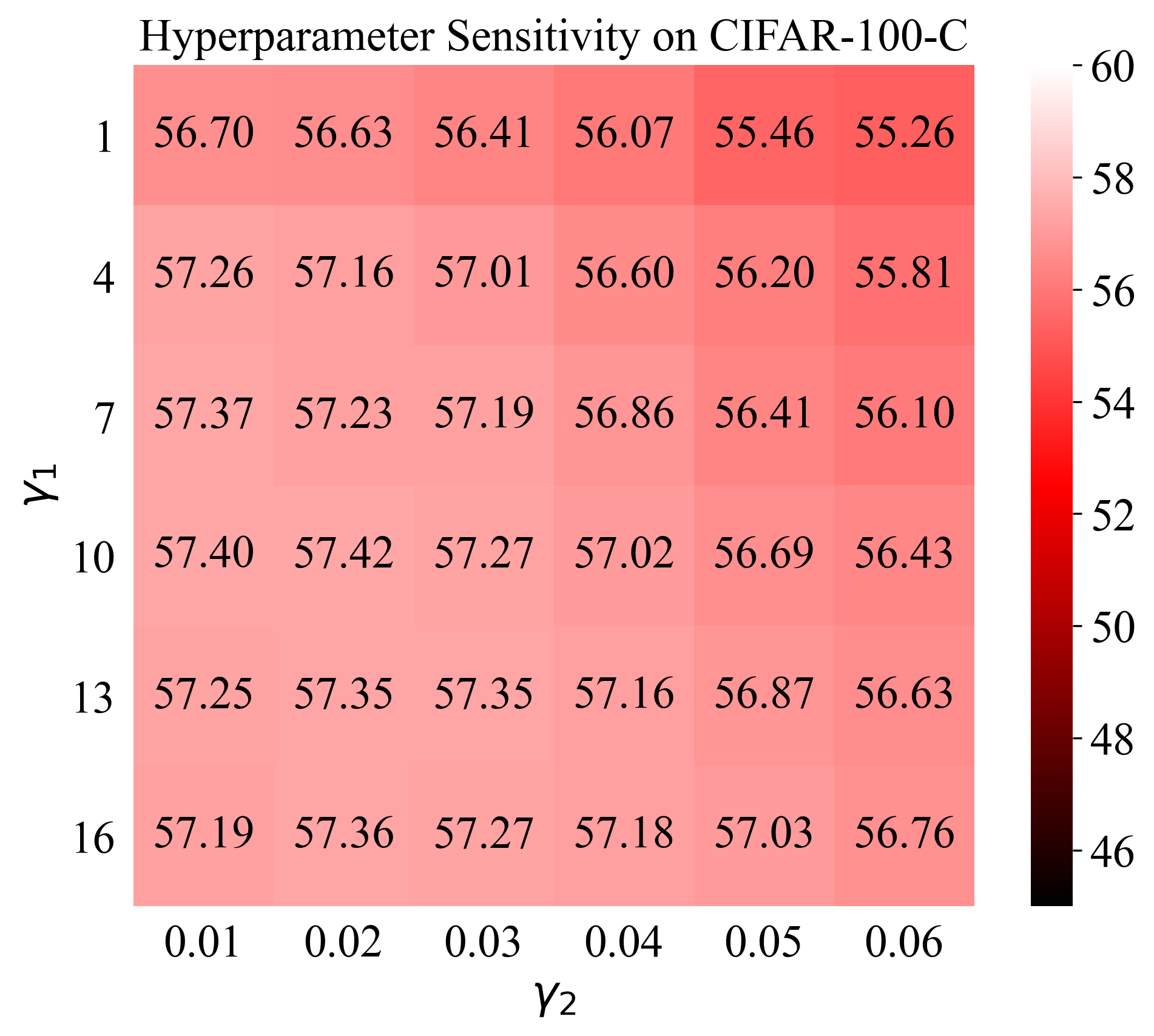}
        \vspace{-8pt}
        \label{fig:heatmapb}
    \end{subfigure}
    \captionsetup{skip=0pt}
    \caption{OSCR heatmaps of ROSETTA with TENT adaptation across hyperparameter settings on CIFAR-10-C and CIFAR-100-C, demonstrating consistent robustness under varying configurations.}
    \label{fig:heatmap}
\end{figure}

\noindent\textbf{Hyperparameter Sensitivity.}
We conduct sensitivity analyses on the hyperparameters $\gamma_1$ and $\gamma_2$, as shown in Fig. \ref{fig:heatmap}, using OSCR as an overall metric to evaluate model performance. 
In our ROSETTA loss, the numerical scales of the terms differ, with the $l_1$-norm values exceeding those of $\mathcal{L}_{\text{ang}}$, which range from 0 to 1.
Therefore, $\gamma_2$ is generally smaller than $\gamma_1$ to balance these two losses. Consistent with UniEnt, we employ a smaller value of $\gamma_2$ for CIFAR-100-C than CIFAR-10-C. This scale discrepancy also varies across datasets, as different category granularity and feature dimensionality lead to distinct feature-norm magnitudes. Experimental results reveal that the OSCR gap between the best and worst values is 2.18 for CIFAR-10-C and 1.9 for CIFAR-100-C, 
demonstrating that our method is robust to variations in $\gamma_1$ and $\gamma_2$.
\vspace{0.1cm}

% \begin{figure}[t]
%     \centering
%     % 第一行
%     \begin{subfigure}[b]{0.495\linewidth}
%         \centering
%         \includegraphics[width=1\linewidth]{image/heatmap_cifar10.png}  % 替换为你的图片路径
%         % \caption{}
%         \vspace{-9pt}
%         \label{fig:heatmapa}
%     \end{subfigure}
%     \hfill
%     \begin{subfigure}[b]{0.495\linewidth}
%     \centering
%     \includegraphics[width=1\linewidth]{image/heatmap_cifar100.png}  % 替换为你的图片路径
%     % \caption{}
%     \vspace{-9pt}
%     \label{fig:heatmapb}
%     \end{subfigure}
%     \caption{Heatmap showing OSCR performance of ROSETTA with TENT adaptation across hyperparameter settings on CIFAR-10-C and CIFAR-100-C. 
%     % Color ranges from white to red to black, with higher OSCR values indicating better open-set recognition. 
%     % (a) CIFAR10-C results; (b) CIFAR100-C results, 
%     Demonstrating ROSETTA's robustness across different settings.}
%     % \vspace{-2pt}
%     \label{fig:heatmap}
% \end{figure}
% \vspace{0.1cm}

% \begin{figure}% 表示强制放置在当前位置
%     \centering
%     \includegraphics[width=0.8\linewidth]{image/ood_gradually.png}  % 替换为你图片的文件名
%     \caption{Impact of scaled OOD loss weight with $\tau$ on ID classification and OOD detection. As OOD loss weight increases, UniEnt and UniEnt+ show improved OOD detection but compromised ID classification, while our method maintains superior performance in both.}  % 图片标题
%     \label{fig:weight of ood loss}  % 设置标签，用于引用
% \end{figure}

\vspace{3pt}  % 你可以调节上下间距

\begin{figure}[t]
    \centering
    \includegraphics[width=0.95\linewidth]{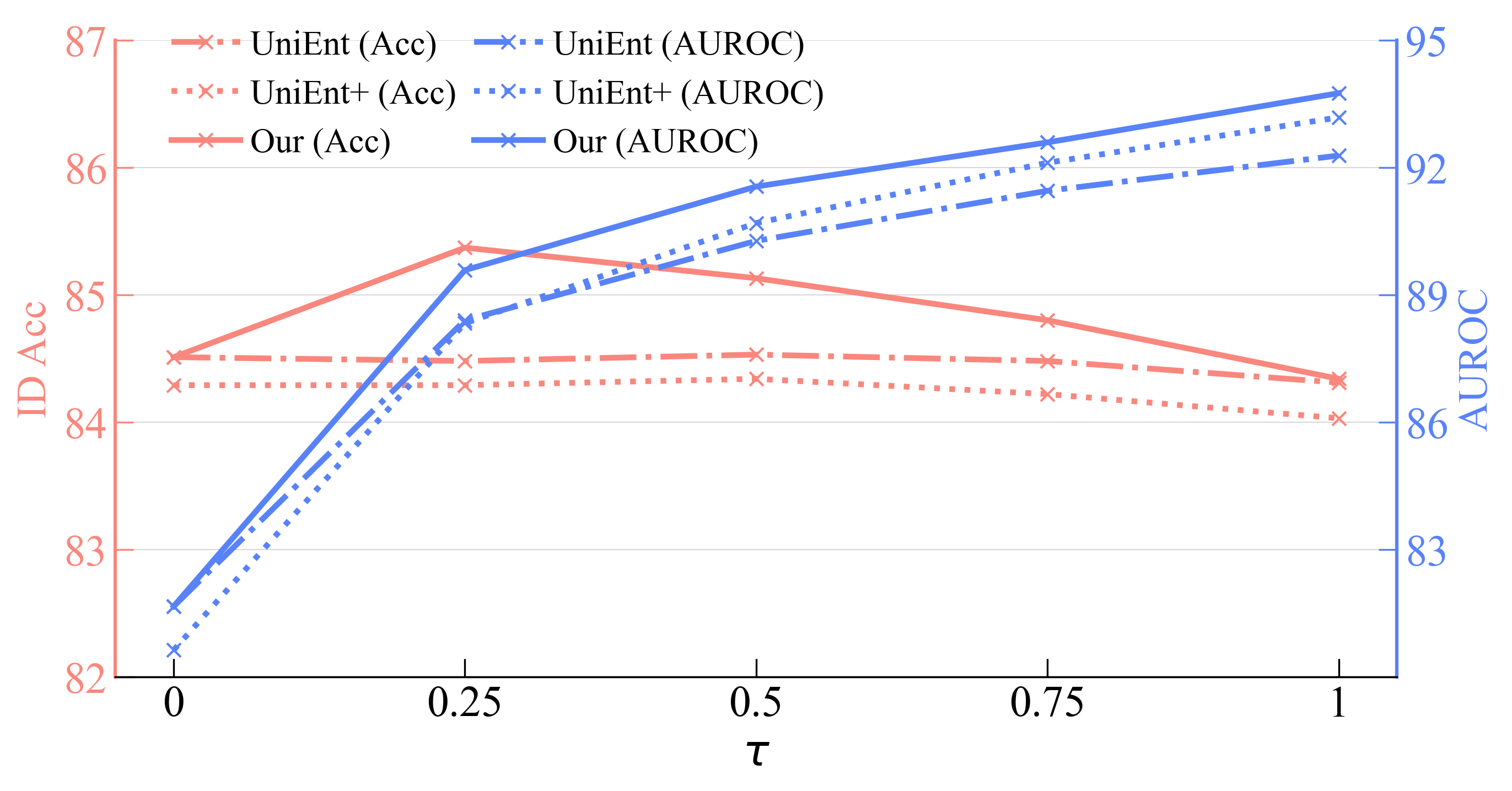}
    \captionsetup{skip=0pt}
    \caption{Effect of increasing OOD loss weight $\tau$ on ID classification and OOD detection. Our method achieves balanced and superior performance on both tasks.}
    \label{fig:weight of ood loss}
\end{figure}

% \noindent\textbf{Impact of csOOD Detection Loss on Model Performance.}
\noindent\textbf{Effect on csID Classification.}
To highlight that our method has minimal impact on csID classification performance, we conduct an experiment examining model performance under varying weights of the OOD detection loss.
% To demonstrate that our method minimally affects csID classification, we evaluate model performance with different OOD detection loss weights.
For UniEnt and UniEnt+, we set the new $\beta_2$ as $\beta_2 \cdot \tau$. For our method, we adjust both $\gamma_2$ and $\gamma_3$ to $\gamma_2 \cdot \tau$ and $\gamma_3 \cdot \tau$, respectively. The scaling factor $\tau$ is gradually increased from 0 to 1, allowing us to analyze the effects of progressively increasing the csOOD detection loss weight on both csID classification and csOOD detection. As shown in Fig. \ref{fig:weight of ood loss}, the results indicate that as the csOOD loss weight increases, OOD AUROC improves, while ID classification accuracy declines for UniEnt and UniEnt+. In contrast, our method maintains ID accuracy as the OOD loss weight increases, ultimately achieving superior performance over UniEnt and UniEnt+ in both ID classification and OOD detection.

\section{Summary}

We analyze the underlying mechanisms of loss functions used in current open-set test-time adaptation methods and 
identify their limitations in simultaneously maintaining both high csID accuracy and effective csOOD detection performance. To address these issues, we propose the ROSETTA loss, which decouples the loss terms for csID classification and csOOD detection. ROSETTA consists of an angular loss and a feature norm loss, both promoting a more uniform logits distribution for csOOD samples, thereby facilitating improved OOD detection. ROSETTA can also be integrated with other entropy-based test-time adaptation methods to enhance their effectiveness in open-set TTA.

\noindent\textbf{Limitations.} Our method is designed to enhance open-set TTA approaches that rely on entropy maximization. It may not be directly applicable to methods based on other techniques. 
% Future work can explore the design of a more universal approach.
Future work may explore more general methods beyond entropy maximization.

%\textbf{Societal Impact.} Our method performs well under both noisy conditions and novel object scenarios, which is critical for safety-sensitive domains such as autonomous driving and robotics. It contributes to more reliable and informed decision-making in real-world environments.


{
    \small
    \bibliographystyle{ieeenat_fullname}
    \bibliography{main}

@String(CVPR= {IEEE Conf. Comput. Vis. Pattern Recog.})

@String(ECCV= {Eur. Conf. Comput. Vis.})

@String(NIPS= {Adv. Neural Inform. Process. Syst.})

@String(BMVC= {Brit. Mach. Vis. Conf.})

@String(ICLR = {Int. Conf. Learn. Represent.})

@String(CVPR  = {CVPR})

@String(ECCV  = {ECCV})

@String(NIPS  = {NeurIPS})

@String(BMVC  =	{BMVC})

@String(ICLR  = {ICLR})

@article{khurana2021sita,
  title={Sita: Single image test-time adaptation},
  author={Khurana, Ansh and Paul, Sujoy and Rai, Piyush and Biswas, Soma and Aggarwal, Gaurav},
  journal={arXiv preprint arXiv:2112.02355},
  year={2021}
}

@article{nado2020evaluating,
  title={Evaluating prediction-time batch normalization for robustness under covariate shift},
  author={Nado, Zachary and Padhy, Shreyas and Sculley, D and D'Amour, Alexander and Lakshminarayanan, Balaji and Snoek, Jasper},
  journal={arXiv preprint arXiv:2006.10963},
  year={2020}
}

@inproceedings{niu2022efficient,
  title={Efficient test-time model adaptation without forgetting},
  author={Niu, Shuaicheng and Wu, Jiaxiang and Zhang, Yifan and Chen, Yaofo and Zheng, Shijian and Zhao, Peilin and Tan, Mingkui},
  booktitle={International conference on machine learning},
  pages={16888--16905},
  year={2022},
  organization={PMLR}
}

@article{wang2020tent,
  title={Tent: Fully test-time adaptation by entropy minimization},
  author={Wang, Dequan and Shelhamer, Evan and Liu, Shaoteng and Olshausen, Bruno and Darrell, Trevor},
  journal={arXiv preprint arXiv:2006.10726},
  year={2020}
}

@inproceedings{wang2022continual,
  title={Continual test-time domain adaptation},
  author={Wang, Qin and Fink, Olga and Van Gool, Luc and Dai, Dengxin},
  booktitle={Proceedings of the IEEE/CVF Conference on Computer Vision and Pattern Recognition},
  pages={7201--7211},
  year={2022}
}

@inproceedings{yuan2023robust,
  title={Robust test-time adaptation in dynamic scenarios},
  author={Yuan, Longhui and Xie, Binhui and Li, Shuang},
  booktitle={Proceedings of the IEEE/CVF Conference on Computer Vision and Pattern Recognition},
  pages={15922--15932},
  year={2023}
}

@article{hendrycks2019benchmarking,
  title={Benchmarking neural network robustness to common corruptions and perturbations},
  author={Hendrycks, Dan and Dietterich, Thomas},
  journal={arXiv preprint arXiv:1903.12261},
  year={2019}
}

@inproceedings{netzer2011reading,
  title={Reading digits in natural images with unsupervised feature learning},
  author={Netzer, Yuval and Wang, Tao and Coates, Adam and Bissacco, Alessandro and Wu, Baolin and Ng, Andrew Y and others},
  booktitle={NIPS workshop on deep learning and unsupervised feature learning},
  volume={2011},
  number={2},
  pages={4},
  year={2011},
  organization={Granada}
}

@article{lim2023ttn,
  title={TTN: A domain-shift aware batch normalization in test-time adaptation},
  author={Lim, Hyesu and Kim, Byeonggeun and Choo, Jaegul and Choi, Sungha},
  journal={arXiv preprint arXiv:2302.05155},
  year={2023}
}

@article{zhou2022domain,
  title={Domain generalization: A survey},
  author={Zhou, Kaiyang and Liu, Ziwei and Qiao, Yu and Xiang, Tao and Loy, Chen Change},
  journal={IEEE Transactions on Pattern Analysis and Machine Intelligence},
  volume={45},
  number={4},
  pages={4396--4415},
  year={2022},
  publisher={IEEE}
}

@article{yang2023fsood,
  title={Full-spectrum out-of-distribution detection},
  author={Yang, Jingkang and Zhou, Kaiyang and Liu, Ziwei},
  journal={International Journal of Computer Vision},
  volume={131},
  number={10},
  pages={2607--2622},
  year={2023},
  publisher={Springer}
}

@inproceedings{gao2024unient,
  title={Unified entropy optimization for open-set test-time adaptation},
  author={Gao, Zhengqing and Zhang, Xu-Yao and Liu, Cheng-Lin},
  booktitle={Proceedings of the IEEE/CVF Conference on Computer Vision and Pattern Recognition},
  pages={23975--23984},
  year={2024}
}

@article{dhamija2018reducing,
  title={Reducing network agnostophobia},
  author={Dhamija, Akshay Raj and G{\"u}nther, Manuel and Boult, Terrance},
  journal={Advances in Neural Information Processing Systems},
  volume={31},
  year={2018}
}

@article{yu2020out,
  title={Out-of-distribution detection for reliable face recognition},
  author={Yu, Chang and Zhu, Xiangyu and Lei, Zhen and Li, Stan Z},
  journal={IEEE Signal Processing Letters},
  volume={27},
  pages={710--714},
  year={2020},
  publisher={IEEE}
}

@inproceedings{chen2020norm,
  title={Norm-aware embedding for efficient person search},
  author={Chen, Di and Zhang, Shanshan and Yang, Jian and Schiele, Bernt},
  booktitle={Proceedings of the IEEE/CVF conference on computer vision and pattern recognition},
  pages={12615--12624},
  year={2020}
}

@inproceedings{meng2021magface,
  title={Magface: A universal representation for face recognition and quality assessment},
  author={Meng, Qiang and Zhao, Shichao and Huang, Zhida and Zhou, Feng},
  booktitle={Proceedings of the IEEE/CVF conference on computer vision and pattern recognition},
  pages={14225--14234},
  year={2021}
}

@inproceedings{park2023understanding,
  title={Understanding the feature norm for out-of-distribution detection},
  author={Park, Jaewoo and Chai, Jacky Chen Long and Yoon, Jaeho and Teoh, Andrew Beng Jin},
  booktitle={Proceedings of the IEEE/CVF International Conference on Computer Vision},
  pages={1557--1567},
  year={2023}
}

@article{hendrycks2020augmix,
  title={{AugMix}: A Simple Data Processing Method to Improve Robustness and Uncertainty},
  author={Hendrycks, Dan and Mu, Norman and Cubuk, Ekin D. and Zoph, Barret and Gilmer, Justin and Lakshminarayanan, Balaji},
  journal={Proceedings of the International Conference on Learning Representations (ICLR)},
  year={2020}
}

@inproceedings{croce2021robustbench,
  title     = {RobustBench: a standardized adversarial robustness benchmark},
  author    = {Croce, Francesco and Andriushchenko, Maksym and Sehwag, Vikash and Debenedetti, Edoardo and Flammarion, Nicolas and Chiang, Mung and Mittal, Prateek and Matthias Hein},
  booktitle = {Thirty-fifth Conference on Neural Information Processing Systems Datasets and Benchmarks Track},
  year      = {2021},
  url       = {https://openreview.net/forum?id=SSKZPJCt7B}
}

@inproceedings{DBLP:conf/bmvc/ZagoruykoK16,
  author       = {Sergey Zagoruyko and
                  Nikos Komodakis},
  editor       = {Richard C. Wilson and
                  Edwin R. Hancock and
                  William A. P. Smith},
  title        = {Wide Residual Networks},
  booktitle    = {Proceedings of the British Machine Vision Conference 2016, {BMVC}
                  2016, York, UK, September 19-22, 2016},
  publisher    = {{BMVA} Press},
  year         = {2016},
  url          = {https://bmva-archive.org.uk/bmvc/2016/papers/paper087/index.html},
  bibsource    = {dblp computer science bibliography, https://dblp.org}
}

@inproceedings{chen2022contrastive,
  title={Contrastive test-time adaptation},
  author={Chen, Dian and Wang, Dequan and Darrell, Trevor and Ebrahimi, Sayna},
  booktitle={Proceedings of the IEEE/CVF Conference on Computer Vision and Pattern Recognition},
  pages={295--305},
  year={2022}
}

@inproceedings{choi2022improving,
  title={Improving test-time adaptation via shift-agnostic weight regularization and nearest source prototypes},
  author={Choi, Sungha and Yang, Seunghan and Choi, Seokeon and Yun, Sungrack},
  booktitle={European Conference on Computer Vision},
  pages={440--458},
  year={2022},
  organization={Springer}
}

@inproceedings{liang2020we,
  title={Do we really need to access the source data? source hypothesis transfer for unsupervised domain adaptation},
  author={Liang, Jian and Hu, Dapeng and Feng, Jiashi},
  booktitle={International conference on machine learning},
  pages={6028--6039},
  year={2020},
  organization={PMLR}
}

@inproceedings{hendrycks2021natural,
  title={Natural adversarial examples},
  author={Hendrycks, Dan and Zhao, Kevin and Basart, Steven and Steinhardt, Jacob and Song, Dawn},
  booktitle={Proceedings of the IEEE/CVF conference on computer vision and pattern recognition},
  pages={15262--15271},
  year={2021}
}

@inproceedings{lee2023towards,
  title={Towards open-set test-time adaptation utilizing the wisdom of crowds in entropy minimization},
  author={Lee, Jungsoo and Das, Debasmit and Choo, Jaegul and Choi, Sungha},
  booktitle={Proceedings of the IEEE/CVF International Conference on Computer Vision},
  pages={16380--16389},
  year={2023}
}

@article{iwasawa2021test,
  title={Test-time classifier adjustment module for model-agnostic domain generalization},
  author={Iwasawa, Yusuke and Matsuo, Yutaka},
  journal={Advances in Neural Information Processing Systems},
  volume={34},
  pages={2427--2440},
  year={2021}
}

@article{van2008visualizing,
  title={Visualizing data using t-SNE.},
  author={Van der Maaten, Laurens and Hinton, Geoffrey},
  journal={Journal of machine learning research},
  volume={9},
  number={11},
  year={2008}
}

@inproceedings{li2023robustness,
  title={On the robustness of open-world test-time training: Self-training with dynamic prototype expansion},
  author={Li, Yushu and Xu, Xun and Su, Yongyi and Jia, Kui},
  booktitle={Proceedings of the IEEE/CVF International Conference on Computer Vision},
  pages={11836--11846},
  year={2023}
}

@inproceedings{yu2024stamp,
  title={STAMP: Outlier-Aware Test-Time Adaptation with Stable Memory Replay},
  author={Yu, Yongcan and Sheng, Lijun and He, Ran and Liang, Jian},
  booktitle={European Conference on Computer Vision},
  pages={375--392},
  year={2024},
  organization={Springer}
}

@article{xie2021segformer,
  title={SegFormer: Simple and efficient design for semantic segmentation with transformers},
  author={Xie, Enze and Wang, Wenhai and Yu, Zhiding and Anandkumar, Anima and Alvarez, Jose M and Luo, Ping},
  journal={Advances in neural information processing systems},
  volume={34},
  pages={12077--12090},
  year={2021}
}

@inproceedings{Cordts2016Cityscapes,
title={The Cityscapes Dataset for Semantic Urban Scene Understanding},
author={Cordts, Marius and Omran, Mohamed and Ramos, Sebastian and Rehfeld, Timo and Enzweiler, Markus and Benenson, Rodrigo and Franke, Uwe and Roth, Stefan and Schiele, Bernt},
booktitle={Proc. of the IEEE Conference on Computer Vision and Pattern Recognition (CVPR)},
year={2016}
}

@inproceedings{sakaridis2021acdc,
  title={ACDC: The adverse conditions dataset with correspondences for semantic driving scene understanding},
  author={Sakaridis, Christos and Dai, Dengxin and Van Gool, Luc},
  booktitle={Proceedings of the IEEE/CVF international conference on computer vision},
  pages={10765--10775},
  year={2021}
}

@article{dong2025towards,
  title={Towards robust multimodal open-set test-time adaptation via adaptive entropy-aware optimization},
  author={Dong, Hao and Chatzi, Eleni and Fink, Olga},
  journal={arXiv preprint arXiv:2501.13924},
  year={2025}
}

@inproceedings{chan2021entropy,
  title={Entropy maximization and meta classification for out-of-distribution detection in semantic segmentation},
  author={Chan, Robin and Rottmann, Matthias and Gottschalk, Hanno},
  booktitle={Proceedings of the ieee/cvf international conference on computer vision},
  pages={5128--5137},
  year={2021}
}

@inproceedings{zhao2024segment,
  title={Segment every out-of-distribution object},
  author={Zhao, Wenjie and Li, Jia and Dong, Xin and Xiang, Yu and Guo, Yunhui},
  booktitle={Proceedings of the IEEE/CVF Conference on Computer Vision and Pattern Recognition},
  pages={3910--3920},
  year={2024}
}

@article{de2025deep,
  title={Deep Out-of-Distribution Uncertainty Quantification via Weight Entropy Maximization},
  author={de Mathelin, Antoine and Deheeger, Fran{\c{c}}ois and Mougeot, Mathilde and Vayatis, Nicolas},
  journal={Journal of Machine Learning Research},
  volume={26},
  number={4},
  pages={1--68},
  year={2025}
}

@article{macedo2021enhanced,
  title={Enhanced isotropy maximization loss: Seamless and high-performance out-of-distribution detection simply replacing the softmax loss},
  author={Mac{\^e}do, David and Ludermir, Teresa},
  journal={arXiv preprint arXiv:2105.14399},
  year={2021}
}

@article{hendrycks2016baseline,
  title={A baseline for detecting misclassified and out-of-distribution examples in neural networks},
  author={Hendrycks, Dan and Gimpel, Kevin},
  journal={arXiv preprint arXiv:1610.02136},
  year={2016}
}

@article{dong2023simmmdg,
  title={SimMMDG: A simple and effective framework for multi-modal domain generalization},
  author={Dong, Hao and Nejjar, Ismail and Sun, Han and Chatzi, Eleni and Fink, Olga},
  journal={Advances in Neural Information Processing Systems},
  volume={36},
  pages={78674--78695},
  year={2023}
}

@inproceedings{damen2018scaling,
  title={Scaling egocentric vision: The epic-kitchens dataset},
  author={Damen, Dima and Doughty, Hazel and Farinella, Giovanni Maria and Fidler, Sanja and Furnari, Antonino and Kazakos, Evangelos and Moltisanti, Davide and Munro, Jonathan and Perrett, Toby and Price, Will and others},
  booktitle={Proceedings of the European conference on computer vision (ECCV)},
  pages={720--736},
  year={2018}
}

@article{gardner2006exponential,
  title={Exponential smoothing: The state of the art—Part II},
  author={Gardner Jr, Everette S},
  journal={International journal of forecasting},
  volume={22},
  number={4},
  pages={637--666},
  year={2006},
  publisher={Elsevier}
}

@book{kushner2003stochastic,
  title={Stochastic approximation and recursive algorithms and applications},
  author={Kushner, Harold J and Yin, G George},
  year={2003},
  publisher={Springer}
}

@InProceedings{Liu_2025_CVPR,
    author    = {Liu, Yuhang and Zhao, Wenjie and Guo, Yunhui},
    title     = {H2ST: Hierarchical Two-Sample Tests for Continual Out-of-Distribution Detection},
    booktitle = {Proceedings of the IEEE/CVF Conference on Computer Vision and Pattern Recognition (CVPR)},
    month     = {June},
    year      = {2025},
    pages     = {15413-15423}
}

@article{liang2017enhancing,
  title={Enhancing the reliability of out-of-distribution image detection in neural networks},
  author={Liang, Shiyu and Li, Yixuan and Srikant, Rayadurgam},
  journal={arXiv preprint arXiv:1706.02690},
  year={2017}
}
}


\clearpage
\setcounter{page}{1}
% \maketitlesupplementary

\section{Appendix}
\label{sec:supp}
To further demonstrate the effectiveness and robustness of our method, we perform additional experiments and analyses.
\begin{enumerate}
    \item First, we present the details of the segmentation task in Section \ref{sec:details}.
    
 % and evaluate the performance of the distribution-aware filter used in UniEnt \cite{gao2024unient} in Section \ref{sec:daf}. 
 % Unlike mainstream OOD detection methods, which rely on OOD scores and require carefully selected thresholds for binary predictions, the distribution-aware filter achieves better performance while mitigating the reliance on thresholds. However, it still encounters some misassigned samples. 
    \item Next, we evaluate the robustness of our method under various settings, including imbalanced csID and csOOD samples in Section \ref{subsec:imbalance}, the absence of csOOD samples in Section~\ref{subsec:robustness_absence_csOOD}, and varying numbers of unknown classes within each epoch in Section~\ref{subsec:imbalance OOD class.}.
    \item  We further analyze the use of the $l_1$-norm and an $l_1$-norm-based OOD score NAN \cite{park2023understanding}, as described in Section \ref{norm}. Our findings reveal that the $l_1$-norm for csID samples is initially smaller than that for csOOD samples at the beginning of adaptation, which adversely affects OOD detection performance during the early stages. The underlying mechanism behind this unexpected phenomenon warrants further investigation. Finally, we visualize the feature distribution using  t-SNE in Section~\ref{tsne}.
\end{enumerate}

\subsection{Prototype Update Stability}
\label{appx.Prototype Update Stability}
The exponential moving average (EMA) update used for prototype maintenance is a standard estimator for time-varying means.
Under bounded feature drift and finite variance assumptions, the update is stable and tracks the moving target with a small bias–variance trade-off controlled by the momentum $\alpha$ \cite{kushner2003stochastic, gardner2006exponential}.

\subsection{Implementation Details of Segmentation Task}
\label{sec:details}
In both the Cityscapes and ACDC datasets, pixels are categorized as either classed or void. Classified pixels correspond to ID in Cityscapes and csID in ACDC, while void pixels are treated as inherent OOD, respectively. To integrate our method into a semantic segmentation model, we pass the pixel-wise classification results through the distribution-aware filter by computing the entropy of each pixel. Since the distribution-aware filter produces labels at the pixel level, our method can be seamlessly incorporated into the semantic segmentation adaptation pipeline.

\subsection{Imbalanced csID and csOOD Samples}
\label{subsec:imbalance}
The main experiments use balanced batches of 100 csID and 100 csOOD samples. To assess robustness under class imbalance, we further evaluate on CIFAR-10-C with csOOD-to-csID ratios ranging from 0.2 to 1.0. As reported in Table \ref{table:imbalance1}, our model demonstrates robust performance on imbalanced inputs, achieving better mean performance across all ratios.

\begin{table}[ht]
\centering
\scriptsize
\label{table:performance}
\begin{tabular}{lccccccc}  % <— 普通 tabular, 不再有 extracolsep
\toprule
Method & 0.2 & 0.4 & 0.6 & 0.8 & 1.0 & $\Delta \downarrow$  &Mean\\
\midrule
Source   & 40.00 & 40.03 & 39.98 & 39.92 & 39.87 & \textbf{0.16} &39.96\\
BN Adapt & 49.91 & 49.55 & 48.92 & 47.97 & 47.10 & 2.81 &48.69\\
\midrule
TENT      & 47.68 & 44.12 & 44.06 & 42.90 & 42.16 & 5.52 &44.18\\
+ UniEnt  & 56.84 & 57.48 & 57.13 & 56.77 & 56.26 & 1.22 &56.90\\
+ UniEnt+     & \textbf{57.15} & \textbf{57.59} & 57.24 & 56.88 & 56.33 & 1.26 &57.04\\
\rowcolor{gray!20}
+ Ours        & 56.63 & 57.51 & \textbf{57.51} & \textbf{57.13} & \textbf{56.80}& 0.88 &\textbf{57.12}\\
\bottomrule
\end{tabular}
\vspace{5pt}
\caption{OSCR performance under different csOOD-to-csID imbalance ratios on CIFAR-100-C. The $\Delta$ column indicates the performance variation, calculated as the difference between the maximum and minimum values across the ratios. ROSETTA achieves the best overall accuracy and the most consistent performance across all imbalance settings.}
\label{table:imbalance1}
\end{table}

\subsection{Robustness to the Absence of csOOD Samples}
\label{subsec:robustness_absence_csOOD}
In our setting, we assume the target domain contains two distinct distributions (csID and csOOD). However, csOOD samples may not always be present in practical deployment scenarios. To evaluate the robustness in such cases, we evaluate the model on purely csID target data. As shown in Table~\ref{without csOOD}, our method outperforms the state‑of‑the‑art baselines UniEnt and UniEnt+ on all four benchmarks. We attribute this advantage to eliminating entropy maximization. State‑of‑the‑art methods apply entropy maximization to input without csOOD, driving the model toward uncertain predictions and degrading classification accuracy. By instead optimizing angular and logit‑norm, ROSETTA preserves csID performance when no csOOD data are present and outperforms UniEnt by 2.51\% on ImageNet‑C.

\begin{table}[!htb]
\centering
% 缩小列间距（可选）
\scriptsize
\setlength{\tabcolsep}{3pt}
% 把表格压到 0.8 倍列宽；caption 不受影响
\begin{tabular}{l|cccc}
\toprule
Method & Cifar-10-C & Cifar-100-C & Tiny-ImageNet-C & ImageNet-C\\
\midrule
TENT & \textbf{85.85} & \textbf{62.86} & \textbf{33.13} & \textbf{45.93}\\
+UniEnt  & 83.26 (-2.59) & 60.01 (-2.85) & 36.60 (+3.47) & 44.48 (-1.45)\\
+UniEnt+ & 83.93 (-1.92) & 59.73 (-3.13) & 35.98 (+2.85) & 39.96 (-5.97)\\
\rowcolor{gray!20}
+Ours & \textbf{84.01 (-1.84)} & \textbf{60.39 (-2.47)} & \textbf{36.62 (+3.49)} & \textbf{46.99 (+1.06)}\\
\bottomrule
\end{tabular}
\vspace{5pt}
\caption{Experiments without csOOD data. Our method is more robust than UniEnt and UniEnt+, which alternate in yielding their highest accuracy. Tent occasionally achieves higher Acc since it does not account for csOOD.}
\label{without csOOD}
\end{table}

\subsection{Varying Numbers of Unknown Classes}
\label{subsec:imbalance OOD class.}
To evaluate the robustness of ROSETTA across varying levels of OOD dataset difficulty, we conduct experiments using CIFAR-10-C as csID and SVHN-C as csOOD, with different numbers of unknown classes. The number of classes in SVHN-C ranges from 2 to 10, representing progressively easier to more challenging csOOD datasets. As shown in Table \ref{table:ood class}, ROSETTA demonstrates consistent robustness across different numbers of unknown classes while achieving superior overall performance.

\begin{table}[ht]
\centering
\scriptsize
\label{table:performance}
% 列格式：首列保持 1.8 cm，后面 7 列改为普通居中 c
%\begin{tabular}{@{\hskip 5pt}p{1.8cm}@{\hskip 3pt}*{7}{>{\centering\arraybackslash}c}@{\hskip 5pt}}
\begin{tabular}{lccccccc}  % <— 普通 tabular, 不再有 extracolsep
\toprule
Method & 2 & 4 & 6 & 8 & 10 & $\Delta \downarrow$ & Mean\\
\midrule
Source % \cite{DBLP:conf/bmvc/ZagoruykoK16}   
& 70.84 & 69.28 & 69.32 & 69.18 & 68.44 & 2.40 & 69.41\\
BN Adapt %\cite{nado2020evaluating}          
& 72.56 & 72.48 & 72.52 & 72.44 & 72.14 & 0.42 & 72.43\\
\midrule
TENT %\cite{wang2020tent}                    
& 49.51 & 48.29 & 51.74 & 49.53 & 50.97 & 3.45 & 50.01\\
+ UniEnt %\cite{gao2024unient}              
& 78.71 & 78.39 & 78.28 & 78.13 & 77.82 & 0.89 & 78.27\\
+ UniEnt+ %\cite{gao2024unient}             
& 78.65 & 78.23 & 78.23 & 78.07 & 77.68 & 0.97 & 78.17\\
\rowcolor{gray!20}
+ Ours                                      & \textbf{81.50} & \textbf{81.37} & \textbf{81.84} & \textbf{81.87} & \textbf{81.37} & \textbf{0.50} & \textbf{81.59}\\
\bottomrule
\end{tabular}
\vspace{5pt}
\caption{OSCR performance with increasing numbers of unknown classes in the csOOD set. As the number of unknown classes grows from 2 to 10, the task becomes more challenging. ROSETTA maintains consistently strong results and shows minimal performance degradation, as reflected by the lowest $\Delta$ value.}
\label{table:ood class}
\end{table}

\subsection{Computational Cost of ROSETTA}

To evaluate the computational cost introduced by ROSETTA, we measure the average running time per batch of our method and AEO~\cite{dong2025towards}. The experiments are conducted using HAC as the csID dataset and EPIC-KITCHENS as the OOD dataset, adapting from the human domain to the animal domain. Each experiment is repeated three times, and we report the average results.

The results show that the average running time per batch for AEO is 2.29 ± 0.026 seconds, while our method takes 2.52 ± 0.0047 seconds. This indicates that ROSETTA introduces only a marginal increase in computational cost.

\subsection{Feature Norm as OOD Detection Score} 
\label{norm}
We conducted experiments using $l_1$-norm and NAN \cite{park2023understanding} score as alternatives to the energy score for OOD detection on CIFAR-10-C, CIFAR-100-C and Tiny-ImageNet-C with ROSETTA based on TENT \cite{wang2020tent}, as shown in Table. \ref{table:different score}. Results on CIFAR-10-C suggest that $l_1$-norm is effective for OOD detection, improving FPR95 by 4.62. However, as dataset complexity and model size increase, $l_1$-norm shows decreased performance compared to the energy score, particularly on Tiny-ImageNet-C. 

Our analysis reveal that initially, AUROC for $l_1$-norm-based detection is below 50, as shown in Table \ref{table:each task}, indicating a tendency to misclassify csID as csOOD. Further investigation shows this anomaly may arise from model behavior: while OOD samples are generally expected to have lower $l_1$-norms than ID samples, csID samples initially exhibited lower norms than csOOD samples at the start of adaptation. Consequently, smaller models (e.g., WideResNet) adapt to covariate-shifted data faster than larger models (e.g., ResNet50), resulting in rapid AUROC gains on CIFAR-10-C but slower improvements on Tiny-ImageNet-C. The reason for csID samples having lower feature norms than csOOD samples at the start of adaptation warrants further study.

% \begin{table}[tbp]
%     \scriptsize
%     % \small
%     \centering
%     \begin{tabular*}{\linewidth}{@{\extracolsep{\fill}}rr |cccc@{}}
%         \toprule
%         % \multirow{2}{*}{Method} & \multicolumn{4}{c}{Tiny-ImageNet-C} \\
%         % \cmidrule(lr){3-6}
        
%          \multicolumn{1}{c}{Dataset} & \multicolumn{1}{c|}{OOD Score} & \multicolumn{1}{c}{Acc$\uparrow$} & \multicolumn{1}{c}{AUROC$\uparrow$} & \multicolumn{1}{c}{FPR95$\downarrow$} & \multicolumn{1}{c}{OSCR$\uparrow$} \\
%         \midrule
%         \multirow{3}{*}{CIFAR-10-C} & Energy \cite{liu2020energy} & 84.34 & 93.75 & 29.91 & \textbf{81.37}\\
%           &$l_1$-norm & 84.34 & \textbf{94.01} & \textbf{25.19} & 81.04\\
%           &NAN \cite{park2023understanding}& 84.34 & 92.87 & 35.88 & 80.50 \\
%         \midrule
%         \multirow{3}{*}{CIFAR-100-C} & Energy \cite{liu2020energy}&59.20 & 91.80 & 35.89 & \textbf{56.80}\\
%           &$l_1$-norm & 59.20 & \textbf{92.40} & \textbf{29.89} & 56.01\\
%           &NAN \cite{park2023understanding}& 59.20 & 91.95 & 31.91 & 55.88 \\
%         \midrule
%         \multirow{3}{*}{Tiny-ImageNet-C} & Energy \cite{liu2020energy}& 37.32 & \textbf{64.01} & \textbf{88.53} &\textbf{29.75}\\
%           &$l_1$-norm & 37.32 & 61.54 & 89.01 &27.61\\
%           &NAN \cite{park2023understanding}& 37.32 & 61.45 & 89.83 &27.57 \\
%         \bottomrule
%     \end{tabular*}
%     \caption{OOD detection performance comparison across different OOD scores on CIFAR-10-C, CIFAR-100-C, and Tiny-ImageNet-C datasets. } 
%     \label{table:different score}
% \end{table}

\begin{table}[tbp]
    \scriptsize
    \centering
    % 直接用 tabular，保留列格式
    \begin{tabular}{@{\extracolsep{\fill}}rr|cccc@{}}
        \toprule
        \multicolumn{1}{c}{Dataset} & \multicolumn{1}{c|}{OOD Score} & \multicolumn{1}{c}{Acc$\uparrow$} & \multicolumn{1}{c}{AUROC$\uparrow$} & \multicolumn{1}{c}{FPR95$\downarrow$} & \multicolumn{1}{c}{OSCR$\uparrow$} \\
        \midrule
        \multirow{3}{*}{CIFAR-10-C} & Energy & 84.34 & 93.75 & 29.91 & \textbf{81.37}\\
          & $l_1$-norm                 & 84.34 & \textbf{94.01} & \textbf{25.19} & 81.04\\
           & NAN  & 84.34 & 92.87 & 35.88 & 80.50\\
        \midrule
        \multirow{3}{*}{CIFAR-100-C} & Energy & 59.20 & 91.80 & 35.89 & \textbf{56.80}\\
        & $l_1$-norm                 & 59.20 & \textbf{92.40} & \textbf{29.89} & 56.01\\
         & NAN & 59.20 & 91.95 & 31.91 & 55.88\\
        \midrule
        \multirow{3}{*}{Tiny-ImageNet-C} & Energy & 37.32 & \textbf{64.01} & \textbf{88.53} & \textbf{29.75}\\
    & $l_1$-norm                 & 37.32 & 61.54 & 89.01 & 27.61\\
    & NAN  & 37.32 & 61.45 & 89.83 & 27.57\\
        \bottomrule
    \end{tabular}
    \vspace{5pt}
    \caption{OOD detection performance comparison across different OOD scores on CIFAR-10-C, CIFAR-100-C, and Tiny-ImageNet-C datasets.}
    \label{table:different score}
\end{table}

\begin{table}[!htb]
    \footnotesize
    \centering
    \begin{tabular}{@{\extracolsep{\fill}}r|cccc@{}}
        \toprule
        \multicolumn{1}{c|}{Corruption} & \multicolumn{1}{c}{Acc$\uparrow$} & \multicolumn{1}{c}{AUROC$\uparrow$} & \multicolumn{1}{c}{FPR95$\downarrow$} & \multicolumn{1}{c}{OSCR$\uparrow$} \\
        \midrule
        Gaussian noise & 37.99 & 48.25 & 95.41 & 18.54 \\
        Shot noise & 38.83 & 52.32 & 93.90 & 24.32 \\
        Impulse noise & 30.75 & 51.41 & 94.62 & 20.19 \\
        Defocus blur  & 38.89 & 58.68 & 92.09 & 27.52 \\
        Glass blur    & 28.96 & 58.14 & 91.67 & 20.76 \\
        Motion blur   & 43.60 & 65.47 & 88.92 & 34.01 \\
        Zoom blur     & 45.42 & 65.79 & 88.45 & 35.49 \\
        Snow          & 37.63 & 64.07 & 88.44 & 29.12 \\
        Frost         & 39.66 & 64.41 & 86.94 & 30.53 \\
        Fog           & 33.68 & 61.65 & 89.62 & 25.69 \\
        Brightness    & 41.78 & 66.21 & 86.55 & 33.23 \\
        Contrast      & 13.11 & 57.79 & 89.30 & 8.90  \\
        Elastic       & 39.17 & 66.30 & 86.10 & 30.85 \\
        Pixelate      & 45.06 & 71.57 & 81.01 & 37.60 \\
        JPEG          & 45.26 & 71.07 & 82.11 & 37.46 \\
        \midrule
        Mean          & 37.32 & 61.54 & 89.01 & 27.61 \\
        \bottomrule
    \end{tabular}
    \vspace{5pt}
    \caption{Performance metrics for each task using $l_1$-norm as the OOD score of our method on Tiny-ImageNet-C, ordered by task sequence during adaptation. An AUROC below 50 indicates a tendency to classify OOD samples as ID more frequently than random guessing.}
    \label{table:each task}
\end{table}

\subsection{T-SNE Visualization}
\label{tsne}
% To provide a more comprehensive understanding of the feature space, we present t-SNE visualizations in Fig. \ref{fig:ltsne1}, \ref{fig:ltsne2}, \ref{fig:ltsne3}, \ref{fig:ltsne4}. These visualizations enable a detailed examination of the clustering patterns and separability of features across different classes, highlighting the distinctiveness of the learned representations. The expanded views provide greater clarity for analyzing the relationships between in-distribution and out-of-distribution samples.

\begin{figure*}[!t]  % 改为 figure*
    \centering
    % 第一行
    \begin{subfigure}[b]{0.495\linewidth}
        \centering
        \includegraphics[width=1\linewidth]{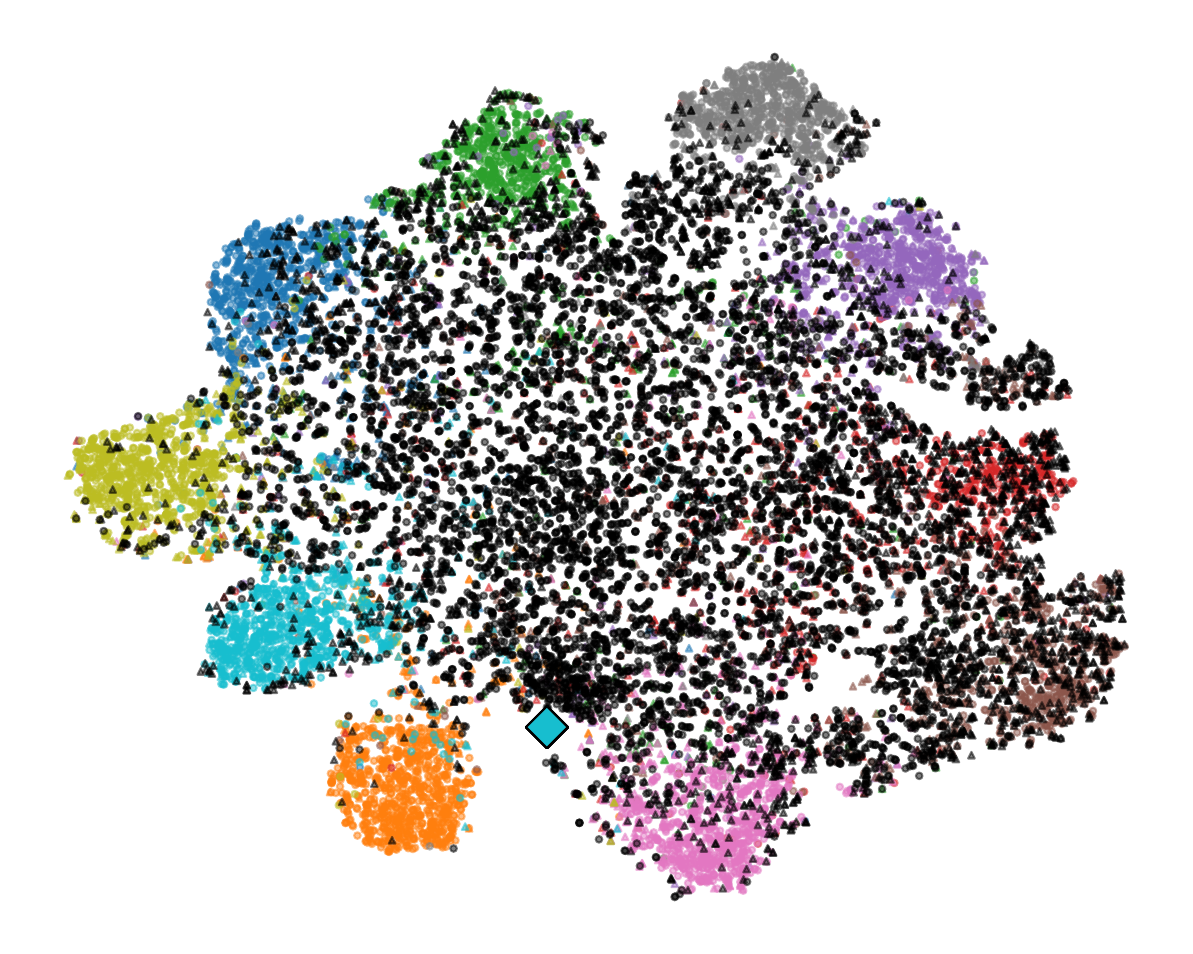}
        \caption{TENT [44] w/ $\mathcal{L}_{t,\text{csID}}$}
        \label{fig:tsnea}
    \end{subfigure}
    \hfill
    \begin{subfigure}[b]{0.495\linewidth}
        \centering
        \includegraphics[width=1\linewidth]{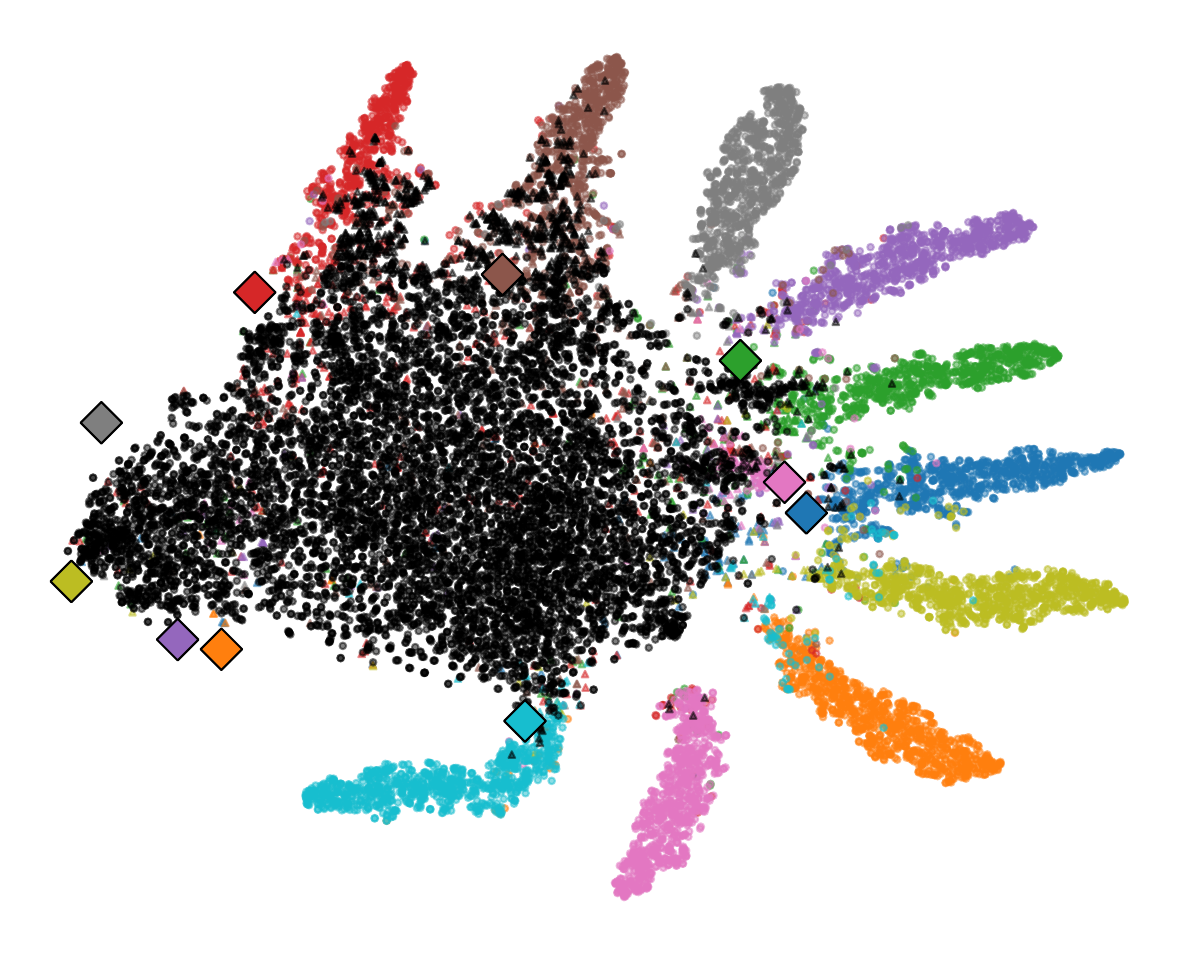}
        \caption{TENT [44] w/ UniEnt+}
        \label{fig:tsneb}
    \end{subfigure}

    \begin{subfigure}[b]{0.495\linewidth}
        \centering
        \includegraphics[width=1\linewidth]{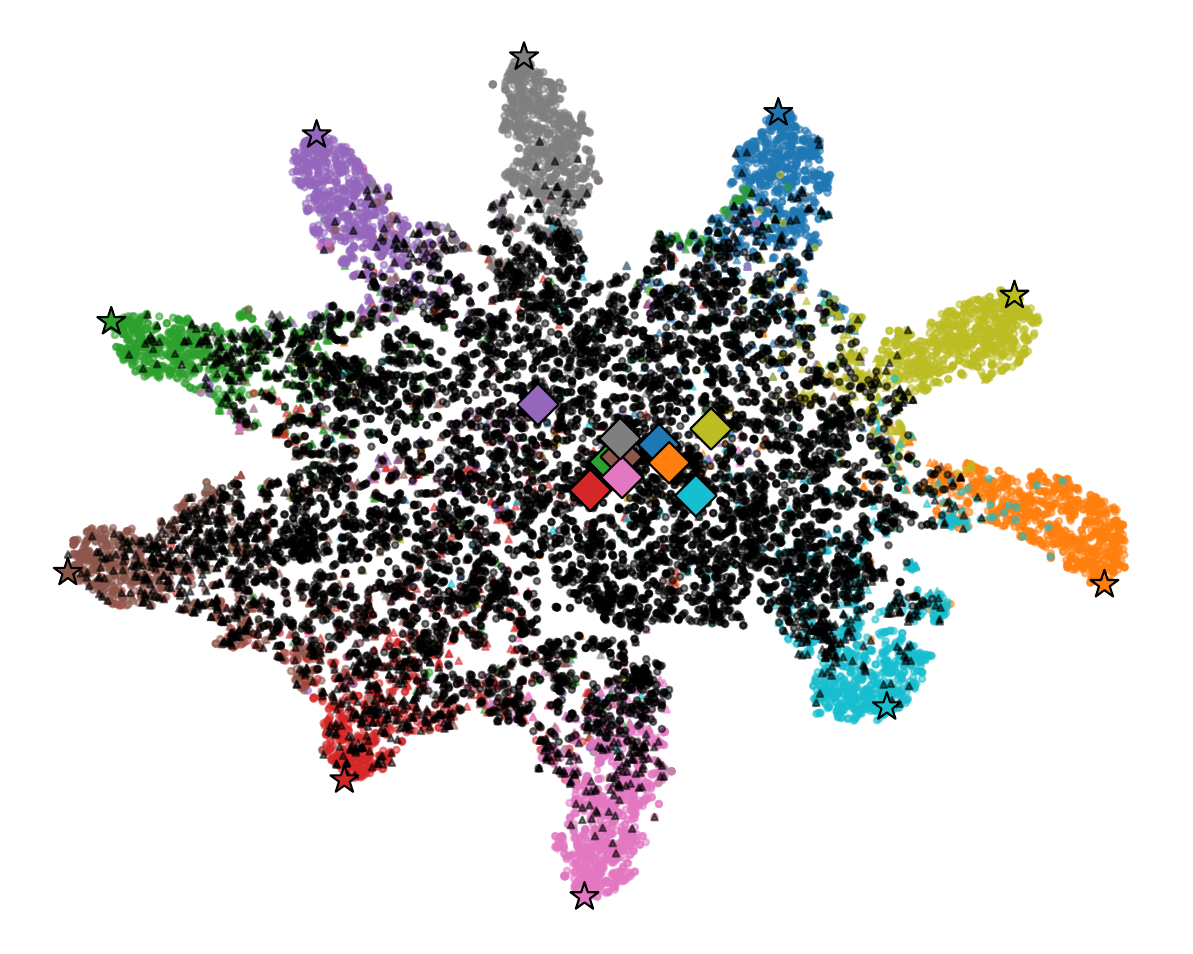}
        \caption{TENT [44] w/ $\mathcal{L}_{t,\text{csID}} + \mathcal{L}_{t,\text{ang}}$}
        \label{fig:tsnec}
    \end{subfigure}
    \hfill
    \begin{subfigure}[b]{0.495\linewidth}
        \centering
        \includegraphics[width=1\linewidth]{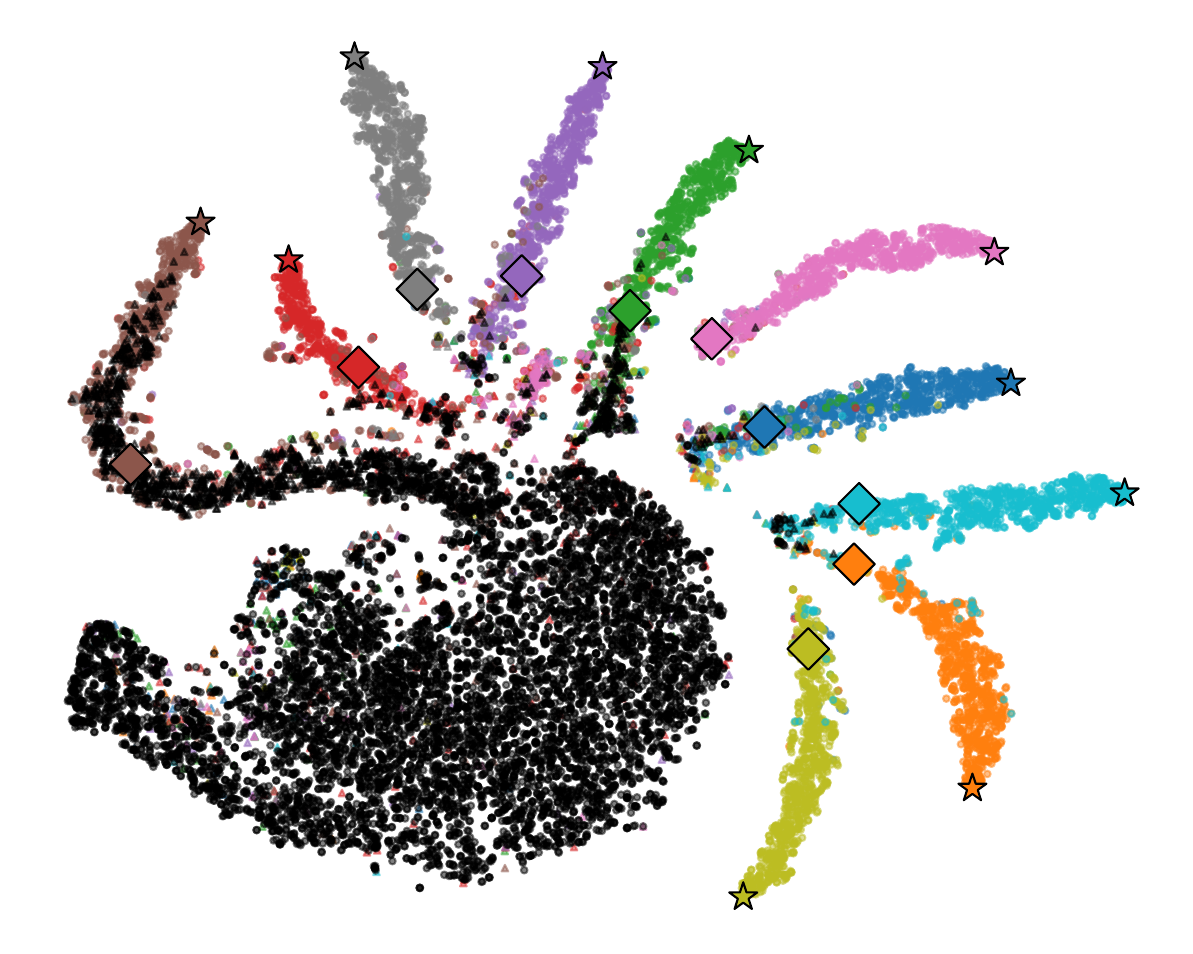}
        \caption{TENT [44] w/ $\mathcal{L}_{t,\text{ROSETTA}}$}
        \label{fig:tsned}
    \end{subfigure}
    \caption{T-SNE Visualization on CIFAR-10-C Test Set with SVHN-C as csOOD. Black markers represent csOOD samples, while colored markers represent csID samples. Circles indicate correctly classified csID samples or correctly detected csOOD samples, whereas triangles denote misclassified csID samples or undetected csOOD samples. Stars represent the embeddings of the updated class prototypes, and diamonds represent the fixed weights of the linear classification layer.}
    \label{fig:tsne_visualization}
\end{figure*}

To further illustrate the effectiveness of our loss function, we visualize the feature vectors of CIFAR-10-C test samples with SVHN-C test samples as csOOD using t-SNE \cite{van2008visualizing}, as shown in Fig. \ref{fig:tsnea}, \ref{fig:tsneb}, \ref{fig:tsnec}, \ref{fig:tsned}. In the t-SNE plot, black triangles indicate misclassified csOOD samples, black circles represent correctly detected csOOD samples, colored triangles denote misclassified csID samples, and colored circles represent correctly classified csID samples. Colored stars mark the embeddings of the updated class prototypes, and colored diamonds represent the fixed weights of the linear classification layer for each class. 

In Fig. \ref{fig:tsnea}, csOOD and csID samples are highly intermixed when only the entropy-minimizing loss is applied, highlighting the difficulty of detecting csOOD samples. By comparing Fig. \ref{fig:tsnea} with Fig. \ref{fig:tsnec}, we observe that the introduction of $\mathcal{L}_{\text{ang}}$ limits feature norm growth, thereby making the fixed classification weights more distinguishable. This addition of $\mathcal{L}_{\text{ang}}$ reduces the norm of csOOD features and increases the cosine similarity between csID samples and their updated class prototypes, thereby improving csOOD detection. Furthermore, when comparing Fig. \ref{fig:tsneb} with Fig. \ref{fig:tsned}, our method achieves a clearer separation between csID and csOOD samples, which is not only beneficial for csID classification but also for csOOD detection in open-set TTA.

% \begin{figure*}[!htbp]
%     \centering
%     \includegraphics[width=\textwidth]{image/12elastic_transformtsne 0 0.pdf}
%     \caption{Enlarged version of the t-SNE visualization originally presented in Fig. \ref{fig:tsnea}. }
%     \label{xx}
% \end{figure*}

% \begin{figure*}[!htbp]
%     \centering
%     \includegraphics[width=\textwidth]{image/10brightnesstsneunient+.pdf}
%     \caption{Enlarged version of the t-SNE visualization originally presented in Fig. \ref{fig:tsneb}
%     }
%     \label{xx}
% \end{figure*}

% \begin{figure*}[!htbp]
%     \centering
%     \includegraphics[width=\textwidth]{image/13pixelatetsne 10 0.pdf}
%     \caption{Enlarged version of the t-SNE visualization originally presented in Fig. \ref{fig:tsnec}
%     }
%     \label{xx}
% \end{figure*}

% \begin{figure*}[!htbp]
%     \centering
%     \includegraphics[width=\textwidth]{image/11contrasttsne.pdf}
%     \caption{Enlarged version of the t-SNE visualization originally presented in Fig. \ref{fig:tsned}
%     }
%     \label{xx}
% \end{figure*}

\clearpage

% \begin{figure*}[p]  % 使用 figure* 来跨越两栏
%     \centering
%     \includegraphics[width=0.8\paperwidth]{image/12elastic_transformtsne 0 0.pdf}  % 使用 paperwidth 而不是 textwidth
%     % \caption{Enlarged version of the t-SNE visualization originally presented in Fig. \ref{fig:tsnea}}
%     \caption{TENT \cite{wang2020tent} w/ $\mathcal{L}_{\text{csID}}$}
% \label{fig:ltsne1}
% \end{figure*}

% % \clearpage  % 强制分页

% \begin{figure*}[!p]
%     \centering
%     \includegraphics[width=0.8\paperwidth]{image/10brightnesstsneunient+.pdf}
%     % \caption{Enlarged version of the t-SNE visualization originally presented in Fig. \ref{fig:tsneb}}
%     \caption{TENT \cite{wang2020tent} w/ UniEnt+}
%     \label{fig:ltsne2}
% \end{figure*}

% \clearpage

% \begin{figure*}[!p]
%     \centering
%     \includegraphics[width=0.8\paperwidth]{image/13pixelatetsne 10 0.pdf}
%     % \caption{Enlarged version of the t-SNE visualization originally presented in Fig. \ref{fig:tsnec}}
%     \caption{TENT \cite{wang2020tent} w/ $\mathcal{L}_{\text{csID}}+ \mathcal{L}_{\text{ang}}$}
%     \label{fig:ltsne3}
% \end{figure*}

% \clearpage

% \begin{figure*}[!p]
%     \centering
%     \includegraphics[width=0.8\paperwidth]{image/11contrasttsne.pdf}
%     % \caption{Enlarged version of the t-SNE visualization originally presented in Fig. \ref{fig:tsned}}
%     \caption{TENT \cite{wang2020tent} w/ $\mathcal{L}_{\text{ROSETTA}}$}
%     \label{fig:ltsne4}
% \end{figure*}

% \clearpage

\end{document}